\pdfoutput=1

\documentclass[12pt]{article}
\usepackage{amsmath}
\usepackage{amssymb}
\usepackage{graphicx}
\usepackage{color}
\usepackage{multirow}
\usepackage[authoryear]{natbib}
\usepackage{rotating}
\usepackage{bbm}
\usepackage{latexsym}
\usepackage{dcolumn}
\usepackage{bm}

\newcommand{\captionfonts}{\footnotesize}

\makeatletter  
\long\def\@makecaption#1#2{%
  \vskip\abovecaptionskip
  \sbox\@tempboxa{{\captionfonts #1: #2}}%
  \ifdim \wd\@tempboxa >\hsize
    {\captionfonts #1: #2\par}
  \else
    \hbox to\hsize{\hfil\box\@tempboxa\hfil}%
  \fi
  \vskip\belowcaptionskip}
\makeatother   

\begin{document}

%
%
%
%

\hspace{13.9cm}

\ \vspace{20mm}\\

{\noindent \LARGE On the complexity of logistic regression models}

\ \\
{\bf \large Nicola Bulso$^{1}$, Matteo Marsili$^{2}$, Yasser Roudi$^{1}$}\\
{\footnotesize $^{\displaystyle 1}$ The Kavli Institute for Systems Neuroscience and Centre for Neural Computation, NTNU, Trondheim, Norway}\\
{\footnotesize $^{\displaystyle 2}$ The Abdus Salam International Centre for Theoretical Physics (ICTP), Trieste, Italy}\\
%

{\noindent \footnotesize {\bf Keywords:} model complexity, logistic regression, Bayesian Model Selection, Minimum Description Length, Jeffreys prior}

\thispagestyle{empty}
\markboth{}{NC instructions}
\ \vspace{-0mm}\\

\begin{center} {\bf Abstract} \end{center}
We investigate the complexity of logistic regression models which is defined by counting the number of indistinguishable distributions that the model can represent \citep{balasubramanian1997statistical}. We find that the complexity of logistic models with binary inputs does not only depend on the number of parameters but also on the distribution of inputs in a non-trivial way which standard treatments of complexity do not address. In particular, we observe that correlations among inputs induce effective dependencies among parameters thus constraining the model and, consequently, reducing its complexity. We derive simple relations for the upper and lower bounds of the complexity. Furthermore, we show analytically that, defining the model parameters on a finite support rather than the entire axis, decreases the complexity in a manner that critically depends on the size of the domain. Based on our findings, we propose a novel model selection criterion which takes into account the entropy of the input distribution. We test our proposal on the problem of selecting the input variables of a logistic regression model in a Bayesian Model Selection framework. In our numerical tests, we find that, while the reconstruction errors of standard model selection approaches (AIC, BIC, $\ell_1$ regularization) strongly depend on the sparsity of the ground truth, the reconstruction error of our method is always close to the minimum in all conditions of sparsity, data size and strength of input correlations.
Finally, we observe that, when considering categorical instead of binary inputs, in a simple and mathematically tractable case, the contribution of the alphabet size to the complexity is very small compared to that of parameter space dimension. We further explore the issue by analysing the dataset of the ``13 keys to the White House'' which is a method for forecasting the outcomes of US presidential elections.

\vspace{\fill}{}


\section{Introduction}
In many studies and in different fields of research, a recurring task is to find the set of explanatory variables, among many candidates, which accurately predict the probability of occurrence of an event. This is the case of identifying risk factors for developing a particular disease, finding the set of neurons in a population which are driving the activity of another neuron or discovering biomarkers from genomic data, to cite a few examples \citep{truett1967multivariate,saeys2007review,hertz2013principles}. When the event we want to predict is binary, such as when estimating the probability of developing a certain disease, a widely employed statistical method is Logistic Regression \citep{cox1958regression}.

Logistic regression model is also interesting because it is the building block of more sophisticated architectures. For instance, under the so called pseudo-likelihood approximation, the difficult task of learning an Ising Model reduces to that of learning many logistic regression models \citep{Besag1972,ravikuumar10,Aurell12}. Another example is that of the Kinetic Ising model \citep{marre2009prediction,Yasser11,Battistin2015}. Both models are popular methods for analysing neural assemblies in computational neuroscience \citep{Schneidman06, roudi2015multi}.

Selecting a set of inputs or predictors\footnote{In the statistics and machine learning literature, the term ``input'' is often used interchangeably with ``predictor'', ``feature'', ``covariate'' or more classically ``independent variable'' \citep{friedman2001elements}. Through the paper we will mainly use the terms ``input'' or ``predictor''.} among some candidates corresponds to comparing different models for the output variable. Thus we search for models with a good trade-off between goodness of fit and model complexity: very simple models might not be able to capture the relevant inputs which are driving the variable we are interested in, whereas complex models are more prone to overfitting \citep{friedman2001elements}. Thus assessing the complexity of a model is an interesting question for model selection \citep[see e.g. ][]{Bulso2016}.

Classical approaches to Model Selection include Bayesian Information Criteria (BIC) \citep{BIC} and Akaike Information Criteria (AIC) \citep{Akaike}. These criteria are derived from different perspectives as approximations of more general quantities in the limit of large samples. However, when the sample is not large enough, these methods might require careful interpretation and application. In this limit, prior information becomes important in the inference process in Bayesian Model Selection. Yet, such information might not be available a priori. In these cases it might be advantageous to use an uninformative prior, such as Jeffreys prior \citep{Jeffreys46}.

As shown in \cite{balasubramanian1997statistical}, the choice of Jeffreys prior in Bayesian Model Selection corresponds to measuring model complexity from a geometric perspective, namely by counting the number of indistinguishable distributions that the model can represent. In this perspective, complex models are models which are able to describe a wide range of probability distributions. In \cite{Myung00}, the authors further show how this approach for characterizing model complexity relates to the Minimum Description Length (MDL) principle \citep{Rissanen1987,Rissanen1996}.

Jeffreys prior is a commonly used prior in Bayesian analysis. It exhibits many nice features among which being uninformative and parametrization invariant \citep{BoxTiao1973} \citep[yet see also][]{kass1996selection}. For Generalized Linear Models, its theoretical properties has been investigated in \cite{ibrahim1991bayesian}. In particular, for binomial data it has been shown that the penalization related to the model complexity under Jeffreys prior depends not only on the number of explanatory variables, which determines the dimensionality of the model, but also on their distribution \citep{chen2008}. Interestingly, Jeffreys prior arises in the large sample size limit also from other perspectives in recently proposed approaches to Model Selection \citep{lamont2017correspondence,mattingly2018maximizing}. In particular, \cite{mattingly2018maximizing} show, among other things, that an optimal prior obtained by maximizing the mutual information between the parameters and their expected data approaches the Jeffreys prior in the limit of abundant data.

In this paper, we address the question of how complex a given logistic model is. We systematically characterize how the complexity depends on the input distribution and prove how the information on model complexity can be used to devise a successful model selection criterion which improves over other well known criteria (BIC, AIC, $\ell_1$ regularization) whose performances instead strongly depend on the sparsity of the data generating model.

The paper is organised as follows. In section \ref{sec:log_reg}, we introduce the notation and define the measure of complexity that we use in the paper. Then we proceed to study, analytically and numerically, the complexity of logistic regression models with binary variables by varying the dimensionality of the model and the correlations among inputs. We also investigate the special case of models whose parameters are defined on a finite support (regularized models). In section \ref{sec:model_selection}, we exploit our results on the complexity to devise a novel model selection criterion and test it on simulated data. Finally in section \ref{sec:degenerate_models}, we investigate the special problem of degenerate models, namely models where parameters might be constrained to being equal. This is an interesting case in model selection since standard recipes based on counting free parameters are not able to distinguish between, for instance, a logistic model with only one parameter multiplying one input and a model with the same parameter multiplying many different inputs. Furthermore, we will see how this issue relates to the problem of evaluating the complexity of categorical variables. We report analytical and numerical results on the complexity along with an application concerning U.S. presidential elections forecasts.

\section{The complexity of logistic regression models with binary inputs}\label{sec:log_reg}
We consider a binary variable $y$ distributed according to a logistic model
\begin{equation}\label{eqn:logistic_regression}
p(y|\bm{x},{\cal{M}}) = \frac{e^{y(b+\bm{w}\cdot\bm{x})}}{2\cosh(b+\bm{w}\cdot\bm{x})},
\end{equation}
with parameters $\bm{\theta} = (b,\bm{w})$ and $N$ binary inputs $\bm{x} = (x_1,x_2,...,x_N)$.
The model is equivalent to a single-layer neural network with $N$ binary input nodes, $\bm{x}$, connected to an output node, $y$, through the weights $\bm{w}$, and bias $b$. Now suppose that we are given $t = 1,.. ,T$ measurements of the inputs, $\hat{\bm{x}} = \{\bm{x}^{(t)}\}_{t=1}^T$, and the output, $\hat{y} = \{y^{(t)}\}_{t=1}^T$, and that the output data has been generated according to the conditional probability distribution in equation \ref{eqn:logistic_regression} for a given choice of the values of parameters (which can also be zero). We are interested in inferring, from the data, the set of non-zero parameters (non-zero components of the vector $\bm{\theta}$). This is a problem of input selection where, given $N$ candidate inputs, we ask which of them are relevant for the variable we want to explain, i.e. which of them is expected to have any influence (non-zero connection) on the output. We proceed following a Bayesian model selection approach. In this framework each possible combination of non-zero parameters represents a model ${\cal M}$ (a set of candidate inputs) and, given $N$ inputs, there are $2^{N+1}$ possible models \footnote{From now on with``models'' we mean non degenerate models, i.e. models whose non-zero parameters are independent of each other, and come back to discuss degenerate models only in the last section.}. The posterior probability of a model ${\cal M}$ given the data $p({\cal M}|\hat{y},\hat{\bm{x}})$, can be expressed by Bayes rule as $p({\cal M}|\hat{y},\hat{\bm{x}}) \propto p(\hat{y}|\hat{\bm{x}},{\cal M}) p({\cal M})$, where $p({\cal M})$ is the prior over models and 
\begin{equation}\label{eqn:bayesian_score}
p(\hat{y}|\hat{\bm{x}},{\cal M}) = \int d\bm\theta\, e^{T\ell (\bm{\theta})} p(\bm{\theta})
\end{equation}
is the probability of measuring $\hat{y}$ given $\hat{\bm{x}}$ and model ${\cal M}$.
In the last equation, $p(\bm{\theta})$ is the prior over parameters and $\ell (\bm{\theta})$ is the normalised log-likelihood function
\begin{equation}\label{eqn:likelihood}
\ell (\bm{\theta}) = \bm{\theta}\cdot\overline{y\bm{x}} - \overline{\log(2\cosh(\bm{\theta}\cdot\bm{x}))},
\end{equation}
where $x_0^{(t)} =1$ and $\theta_0 = b$.
The overbars in equation \ref{eqn:likelihood} stand for averages over all observations, namely $\overline{f(X)} = 1/T\sum_t f(X^{(t)})$ for any function $f(X)$.
Thus, the set of non-zero parameters that has generated the output $\hat{y}$ from the input $\hat{\bm{x}}$ can be inferred by searching for the model that maximizes $p({\cal M}|\hat{y},\hat{\bm{x}})$ among all candidates.
When all models are a priori equally likely, this procedure is equivalent of finding the model that maximizes $p(\hat{y}|\hat{\bm{x}},{\cal M})$, which is given by equation \ref{eqn:bayesian_score}. 

The choice of the prior over parameters represents an important issue when the sample is not very large. In this paper we use Jeffreys prior. Our choice is motivated by the geometrical interpretation of model complexity that originates from it and by its relation with MDL \citep{balasubramanian1997statistical,Myung00}.  Jeffreys prior is defined as follows
\begin{equation}\label{eqn:jeffreys_prior}
J(\bm\theta) = \frac{\sqrt{\det F(\bm\theta)}}{\int d\bm\theta \sqrt{\det F(\bm\theta)}}
\end{equation}
where $F(\bm\theta)$ is the Fisher Information whose elements are $F_{i,j}(\bm{\theta}) = -\text{E}[\partial_{\theta_i,\theta_j}^2 \ell(\bm\theta)]$, where $\text{E}[\cdot]$ is the expected value with respect to the model distribution $p(y|\bm{x},\bm{\theta},{\cal{M}})$. Since the elements of the Hessian matrix $H_{i,j}(\bm{\theta}) =  -\partial_{\theta_i,\theta_j}^2 \ell(\bm\theta)$, calculated from equation \ref{eqn:likelihood}, do not depend on the random variable $y$, it follows that the Fisher Information matrix is simply equal to the Hessian.

With this choice for the prior, the probability $p(\hat{y}|\hat{\bm{x}},{\cal M})$ in equation \ref{eqn:bayesian_score} can be approximated by the following expression \citep[see ][for instance]{balasubramanian1997statistical}:
\begin{equation}\label{eqn:bayesian_approximation}
\log p(\hat{y}|\hat{\bm{x}},{\cal M}) = T\ell(\bm{\theta}^\ast) -\frac{n}{2}\log\frac{T}{2\pi} - \log\int d\bm{\theta} \sqrt{\det F(\bm{\theta})} + O(1/T)
\end{equation}
where $n \leq N$ is the number of non-zero parameters (non-zero components of the parameter vector $\bm\theta$) employed by model $\mathcal{M}$ and $\bm\theta^\ast$ is the maximum likelihood estimate.

In MDL theory the first three terms in equation \ref{eqn:bayesian_approximation}, all together, represent the {\itshape stochastic complexity} of the string $\hat{y}$ given the model $\mathcal{M}$ and the inputs $\hat{x}$, which is the length of the shortest code that can be obtained by encoding the data with the parametric family ${\cal M}$ \citep{Rissanen1987}. The penalty terms in MDL (the second and third term in equation \ref{eqn:bayesian_approximation}), also known as the {\itshape geometric complexity} \citep{Myung00}, are independent of the specific instance of the output data $\hat{y}$ and rather related to intrinsic and geometric properties of the parametric family. Yet it is worth noting that since the Fisher Information in equation \ref{eqn:bayesian_approximation} depends on the input data $\hat{\bm{x}}$, the complexity also depends on the input distribution, as we will discuss thoroughly in the paper.
Finally, the second term in equation \ref{eqn:bayesian_approximation} corresponds to the well known Bayesian Information Criteria \citep{BIC}.

In the following sections we will investigate the typical properties of the geometric complexity for logistic regression models with binary variables (equation \ref{eqn:logistic_regression}). In particular, we will focus on the penalty term
\begin{equation}\label{eqn:def_complexity}
C = \log\int d\bm\theta \sqrt{\det F(\bm\theta)}
\end{equation}
and we will study its dependence on the model size and input distribution. 
For logistic regression models, as defined by equation \ref{eqn:logistic_regression}, the elements of the Fisher Information matrix can be written explicitly in the form 
\begin{equation}\label{eqn:fisher_info}
F_{i,j}(\bm{\theta}) = \sum_{\mu} \nu(\bm{x}^\mu) \cosh^{-2}(\bm{\theta}\cdot\bm{x}^\mu)x^\mu_ix^\mu_j.
\end{equation}
where $\nu(\bm{x}^\mu)$ is the frequency of observing the configuration $\bm{x}^\mu$ in the data, with $\mu = 1, ..., 2^{n}$.
Beyond the interesting theoretical aspects, the complexity represents a higher order term in the expansion of the posterior \ref{eqn:bayesian_approximation} and it is therefore relevant for model selection, particularly when the sample size is not very large. 
From now on we will refer to $C$ in equation \ref{eqn:def_complexity} as the {\itshape complexity}. Notice that we will also use the term complexity to refer to $e^C$ and the difference will be clear from the context.

\subsection{A worked example: n = 2}\label{sec:case_n=2}
We start by considering simple cases which are analytically solvable. The simplest case is when only one parameter is different from zero. When the only non-zero parameter is the bias, i.e $\theta = b$, then the complexity is simply $e^C = \int d\theta \cosh^{-1}(\theta) = \pi$. The same result is retrieved also when the only non-zero parameter is a weight, i.e. $\theta = w$ and the associated input is observed in the configuration $x$ with frequency $\nu(x)$: $e^C = \int d \theta \sqrt{F(\theta)} = \int d\theta \sqrt{\sum_x \nu(x) \cosh^{-2}(\theta x)} = \int d\theta \cosh^{-1}(\theta) = \pi $.

We consider now a model in which only two weights are different from zero, i.e. $\bm\theta = (w_1,w_2)$. The Fisher information is now a two by two matrix whose determinant can be easily calculated: $\det F(\bm{\theta}) = 4 \nu(1-\nu) \cosh^{-2}(\theta_{1} + \theta_{2})\cosh^{-2}(\theta_{1} - \theta_{2})$, where $\nu$ is the frequency that the two inputs, $\bm x = (x_1,x_2)$, have been observed with the same sign. The complexity is therefore given by the integral $\sqrt{4 \nu(1-\nu)} \int d \bm{\theta} \cosh^{-1}(\theta_{1} + \theta_{2})\cosh^{-1}(\theta_{1} - \theta_{2})$ which can be solved by the change of variable, $\phi_1 = \theta_1 + \theta_2$ and $\phi_2 = \theta_1 - \theta_2$, with the result $e^C = \sqrt{\nu(1-\nu)}\pi^2$. 

As it is clear from the above formula, the complexity varies in the range $0\leq e^C \leq \pi^2/2$ according to the degree of correlation between the two inputs. The complexity reaches its maximum, i.e. $\pi^2/2$, when the two inputs are observed with the same sign half of the times, $\nu = 1/2$. It is worth noting that the maximum value for the complexity of the logistic regression model with $n=2$ is smaller than the complexity of a model for two independent outputs, each of which being explained by a logistic regression model with one parameter, which is $\pi^2$. In both cases the models employ two parameters. However, in the latter case, the parameters are completely independent of each other and the volume spanned in the space of distributions will be the maximum achievable given the absence of constraints. Following the same argument, for a generic number of parameters $n$, the complexity of a logistic regression model should always be smaller than $\pi^n$.

In case of strongly correlated inputs, namely when $\nu$ is close to one or zero, the complexity will reach its minimum. In particular, for a sample of size $T$, the smallest non-zero value for $\nu$ is $\nu = 1/T$ and the largest value different from one is $\nu = 1-1/T$. In the latter cases, the complexity attains its lower (non trivial) bound which is equal to $e^C = \pi^2\sqrt{(T-1)/T^2}$. Intuitively, correlations among inputs induce effective dependences among parameters, thus constraining the model into a reduced volume in the space of distributions. In other words, the stronger the correlations, the higher the constraints and, consequently, the lower the complexity. This represents a hard case for model selection. In fact, in the case of two strongly correlated inputs, the three models with at least one non-zero weight parameter multiplying the inputs will have approximatively the same maximum likelihood values. It follows that the selected model will strongly depend on the assumptions pertaining the model selection recipe employed. 

In the limiting case in which $\nu=0$ or $\nu=1$, we get $e^C=0$ and we say that the model is {\itshape redundant}: the model is constrained to a lower dimensional manifold defined by the deterministic dependence between the parameters induced by equal ($x_1 = x_2$) or specular inputs ($x_1 = -x_2$). In this case, the $n$ by $T$ dimensional matrix of input data, $x_{i}^{(t)}$, also called the {\itshape design matrix}, is not full rank. In all cases which are not redundant, i.e. $0< \nu <1$, the matrix is full rank, the complexity is finite and the BIC term is always larger than $C$ in absolute value, consistent with the expansion of equation \ref{eqn:bayesian_approximation} and the geometrical interpretation of the penalization factors as ratio between volumes \citep{balasubramanian1997statistical}.

In the following we derive and discuss some results on the complexity of logistic regression models as a function of the dimension $n$ of the model and of the shape of the input distribution extending the observations drawn from these simple cases.

\subsection{Bounds on geometric complexity}\label{sec:mathematics}
In this section, we will derive and discuss analytical results for the upper and lower bound of the complexity. Additionally, we will support and further investigate the results with numerical simulations.
Without loss of generality, we consider models with only weight parameters, namely $\bm\theta = \bm{w}$ as defined at the beginning of section \ref{sec:log_reg}. In fact, all results derived in this section easily extend also to models with a bias parameter, i.e. $\bm\theta = (b,\bm{w})$, given that the bias term acting on the output can be thought as a weight multiplying a constant input.

The Fisher Information defined in equation \ref{eqn:fisher_info} can be conveniently rewritten as $F(\bm{\theta}) = B^\top B$ by defining the $2^{n}\times n$ matrix $B$ as 
$ B^\mu_i = \sqrt{\nu(\mathbf{x}^\mu) \cosh^{-2}(\bm{\theta}\cdot \mathbf{x}^\mu)} x^\mu_i$,
where we use subscripts and Roman letter for components indices (those ranging from 1 to $n$) whereas superscript and Greek letters for configurations indices (ranging from 1 to $2^{n}$). Moreover the matrix $B$ can be expressed as $B =  DX$, namely as the product of the $2^{n}\times n$ matrix $X$ whose components are given by $x^\mu_i$ and the diagonal matrix $D$ whose elements are $D^{\mu\nu} = \sqrt{\nu(\mathbf{x}^\mu) \cosh^{-2}(\bm{\theta}\cdot \mathbf{x}^\mu)}\delta^{\mu\nu}$.
These mathematical manipulations allow us to exploit Cauchy Binet formula and write the complexity in a more compact form. In fact, by Cauchy Binet formula, the determinant of the Fisher Information reads $\det F(\theta) = \det(B^\top B) = \sum_{\bm\alpha} {\det}^2(B^{\bm\alpha})$,
where $B^{\bm\alpha}$ is now a square matrix obtained from $B$ by selecting $n$ configuration indices over the $2^{n}$ possible ones. The selection is labelled by the vector $\bm\alpha$ and the sum ranges over all ${2^{n}}\choose{n}$ possible $\bm\alpha$ vectors. 
Finally, since $B =  DX$, the determinant of the Fisher Information can be written as $\det F(\theta) = \sum_{\bm\alpha} {\det}^2(X^{\bm\alpha}) \prod_{i = 1}^{n} \nu(\mathbf{x}^{\alpha_i}) \cosh^{-2}(\bm{\theta}\cdot\mathbf{x}^{\alpha_i})$ and consequently the complexity takes the form
\begin{equation}\label{eqn:complexity}
e^{C} =  \int d\bm{\theta} \sqrt{\sum_{\bm\alpha} {\det}^2(X^{\bm\alpha}) \prod_{i = 1}^{n} \nu(\mathbf{x}^{\alpha_i}) \cosh^{-2}(\bm{\theta}\cdot\mathbf{x}^{\alpha_i})}.
\end{equation}
From the above expression we can now find a mathematical approximation for the upper bound of the complexity. By employing the triangular inequality in equation \ref{eqn:complexity}, we move the sum over $\bm\alpha$ outside the square root so that $e^{C}\leq \sum_{\bm\alpha} \det(X^{\bm\alpha}) \prod_{i = 1}^{n} \sqrt{\nu(\bm{x}^{\alpha_i})} I(\mathbf{x}^{\alpha_i})$ where $I(\mathbf{x}^{\alpha_i}) = \int d\bm{\theta} \cosh^{-1}(\bm{\theta}\cdot\mathbf{x}^{\alpha_i})$.
The integral $I(\mathbf{x}^{\alpha_i})$ can be evaluated by making the change of variable $\phi_i =  \bm{\theta}\cdot\mathbf{x}^{\alpha_i}$ and turns out to be equal to $\pi^{n}/\det(X^{\bm\alpha})$. Notice that if ${\det}(X^{\bm\alpha})=0$, there is no contribution to the summation in expression \ref{eqn:complexity}. Therefore the factors $\det(X^{\bm\alpha})$ cancel each other and the sum over $\bm\alpha$ is restricted to those selections $\bm\alpha \in \Omega$ such that $\det(X^{\bm\alpha}) \neq 0$. 
The final result is the following simple inequality
\begin{equation}\label{eqn:math_upbound1}
e^{C}\leq \pi^{n}\sum_{\bm\alpha \in \Omega}\prod_{i = 1}^{n} \sqrt{\nu(\mathbf{x}^{\alpha_i})},
\end{equation}
where the equal sign holds when only $n$ linearly independent configurations are present in the input data matrix (design matrix) as, in this case, there is only one possible selection ${\bm{\alpha}}$ in equation \ref{eqn:complexity}. Notice that equation \ref{eqn:math_upbound1} depends on both the dimensionality $n$ and the input distribution $\nu(\bm{x})$. 

First, we consider the case in which the input distribution is localised on few configurations. In the limiting case in which the distribution is peaked around only one configuration $\xi$, i.e. $\nu(\mathbf{x}^{\mu}) = \delta_{\mu\xi}$, where $\delta$ is the Kronecker delta function (most localised distribution), $e^C$ will be zero. 
As we have already observed in section \ref{sec:case_n=2} for models with $n=2$, in this extreme case the design matrix is not full rank. Consequently the parametrization is redundant: a model with only one parameter would have sufficed to explain the behaviour of the output variable in this particular case. Thus, in general the redundancy is caused by dependencies among the parameters induced by an ``extremely localised" distribution of the input configurations. Those dependencies constrain the model to a lower dimensional manifold whose volume is of zero measure. Whenever this happens, the model represents a redundant description of the reality and one should look for a simpler explanation by reducing the number of parameters and/or combining inputs.

Moreover, it is rather intuitive and easy to see from equation \ref{eqn:math_upbound1} that the parametrization is not redundant if there are at least $n$ linearly independent configurations $\bm{x}^{\mu_i}$ with $i = 1,... ,n$ such that $\nu(\bm{x}^{\mu_i}) \neq 0$ , that is if the design matrix is full rank. In the case of exactly $n$ linearly independent configurations there is only one selection $\bm{\alpha}$ such that $\det X^{\bm{\alpha}}\neq 0$ and equation \ref{eqn:math_upbound1} holds with the equal sign: the result is simply $e^{C} = \pi^{n} \prod_{i = 1}^{n} \sqrt{\nu(\mathbf{x}^{\alpha_i})}$. In this case, given a sample of size $T$, the value of the complexity can vary between two values, $\pi^n \sqrt{(T-n+1)/T^{n}} \leq e^C \leq \pi^n/n^{n/2}$. The upper bound can be derived by maximizing $e^C$ while enforcing the normalisation of the frequencies through a Lagrange multiplier and is achieved when the $n$ linearly independent configurations are observed an equal number of times $\nu(\bm{x}^{\mu_i}) = 1/n$,  $\forall i$. The lower bound can be verified numerically for small values of $n$ and $T$ and corresponds to the situation in which all configurations but one are observed only once, i.e. $\nu(\bm{x}^{\mu_i}) = 1/T$,  $\forall i\neq j$, and $\nu(\bm{x}^{\mu_j}) = (T-n+1)/T$.
The difference between these two values tends to zero as $T$ approaches $n$ from above, while it grows large as $T \gg n$ and $n\gg 1$. In fact, their ratio scales as $(T/n)^{n/2}$. 

\begin{figure}[t!]
\centering
\includegraphics[width = .9\columnwidth]{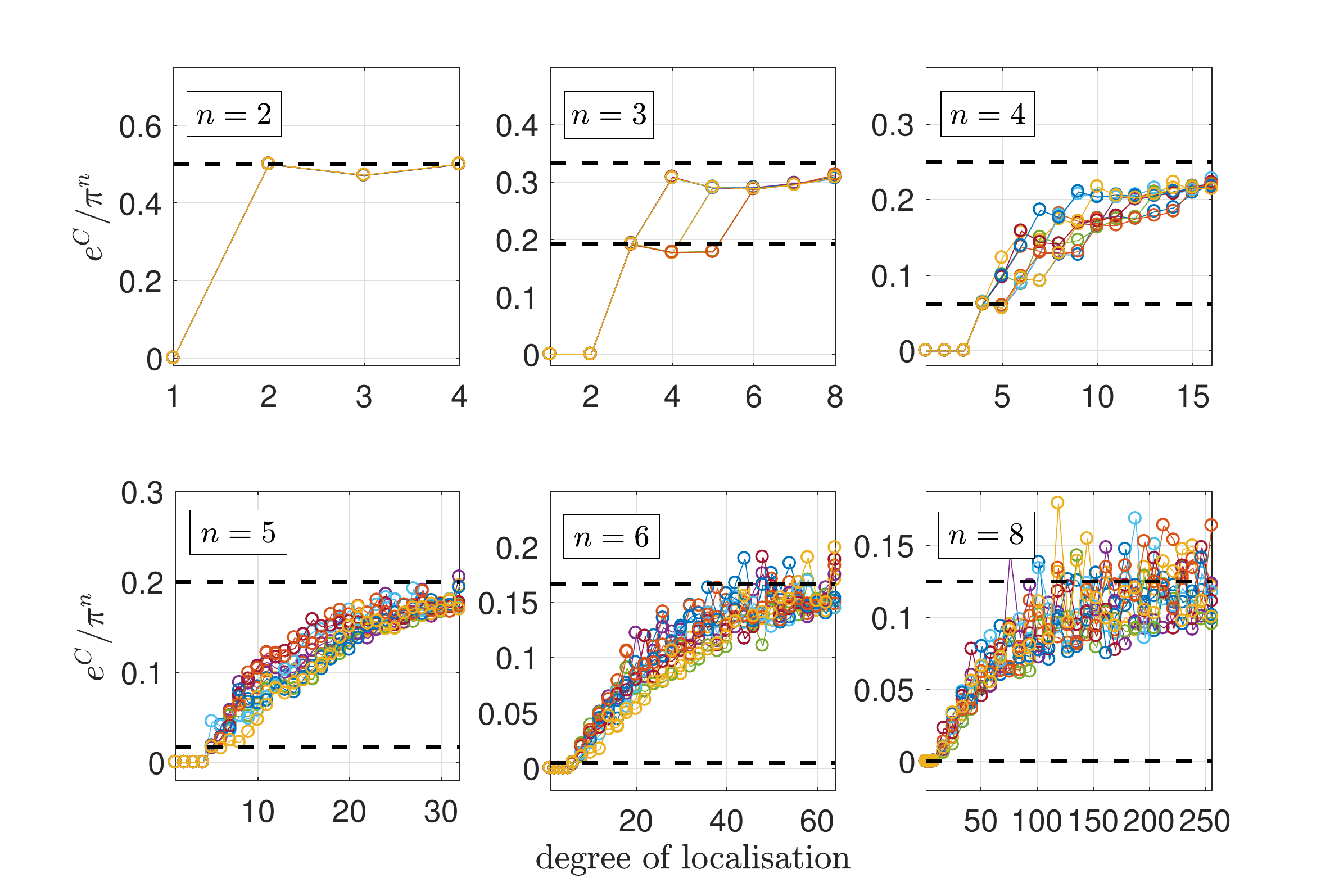}
\caption{The complexity at varying degrees of localisation of the input distribution from $d = 1$ (the inputs are frozen in one configuration) to $d = 2^n$ (the inputs are observed in all possible configurations the same number of times) and for $n = 2, 3, 4, 5, 6, 8$. The values of the complexity are enclosed between two limits: $e^C/\pi^n = n^{-n/2}$ for $d = n$ (bottom dashed line) and  $e^C/\pi^n \sim n^{-1}$ for $d = 2^n$ (top dashed line). For $d <n$ the complexity is zero. The different lines are 10 different random paths: starting from $d = n$ linearly independent configurations and randomly adding configurations until reaching $d=2^n$. For a given value of $d$, the complexity exhibits a high variability which depends on the degree of independence of the selected subset of configurations. Each estimate of the integral in the figure is the result of the average over 20 different Monte Carlo estimates, each of which has been calculated with $10^5$ points. Notice that estimations are less accurate as $n$ grows large due to an increasing variability in the integrand (see Supplemental Information, section 1).}\label{fig:MC_varyingLoc}
\end{figure}

Based on the fact that a more localised input distribution would lead to a redundant parametrization (zero value for the complexity) and given that equation \ref{eqn:math_upbound1} and numerical simulations (see Figure \ref{fig:MC_varyingLoc}) suggests that a more spread distribution would generally lead to an equal or larger value for the complexity, we expect that the lower bound presented above, $e^C = \pi^n \sqrt{(T-n+1)/T^{n}}$, represents the minimum attainable (non zero) value of the complexity. In fact, when $n=2$, it matches the lower bound we found in section \ref{sec:case_n=2}.

Interestingly, in this ``most-localized scenario'', the complexity decreases instead of increasing as the dimensionality of the model increases (it goes at most as $e^C \sim \pi^n/n^{n/2}$). This means that, although adding a new parameter would still increase the dimensionality of the model (larger BIC penalization), it would also result in a more constrained model (decreasing of complexity).

In order to support these results and further investigate the dependence of the complexity on the input distribution, we resort to Monte Carlo simulations for estimating the integral $e^C = \int d\bm\theta \sqrt{\det F(\bm\theta)}$ (details of the simulations are described in Supplemental Information, section 1). 

As a first step, we assume an input distribution which is uniform and different from zero only in a limited number of configurations, $d$. The parameter $d$ tunes the localisation of the input distribution or, in other words, the strength of input correlations. We evaluated the complexity at varying $d$ over the entire range: from $d=1$ to $d = 2^n$. The results are shown in Figure \ref{fig:MC_varyingLoc} for $n=2,3,4,5,6, 8$. 
As predicted, the complexity is zero if less than $n$ configurations have been observed at least once and, when the input distribution is uniformly distributed on exactly $n$ linearly independent configurations, the complexity jumps to $e^C = \pi^n/n^{n/2}$ (bottom black dashed line). For $d>n$, the value of the complexity depends on the degree of independence of the $d$ configurations, namely on the volumes spanned by each subset of $n$ configurations. Therefore for $d>n$ the complexity exhibits a high variability while generally increasing with $d$ from $e^C = \pi^n/n^{n/2}$ toward an upper bound which is achieved when the input distribution is uniform over all possible configurations (top black dashed line). This fact is in line with what has been observed in the case $n=2$ in section \ref{sec:case_n=2} where a uniform distribution leads to the model with the largest complexity. 

Unfortunately, when the distribution is uniform, the inequality \ref{eqn:math_upbound1} provides a trivial upper bound, namely an upper bound larger than expected from simple arguments i.e. $e^C \leq \pi^n$, as argued in \ref{sec:case_n=2}.

\begin{figure}[t!]
\centering
\includegraphics[width = \columnwidth]{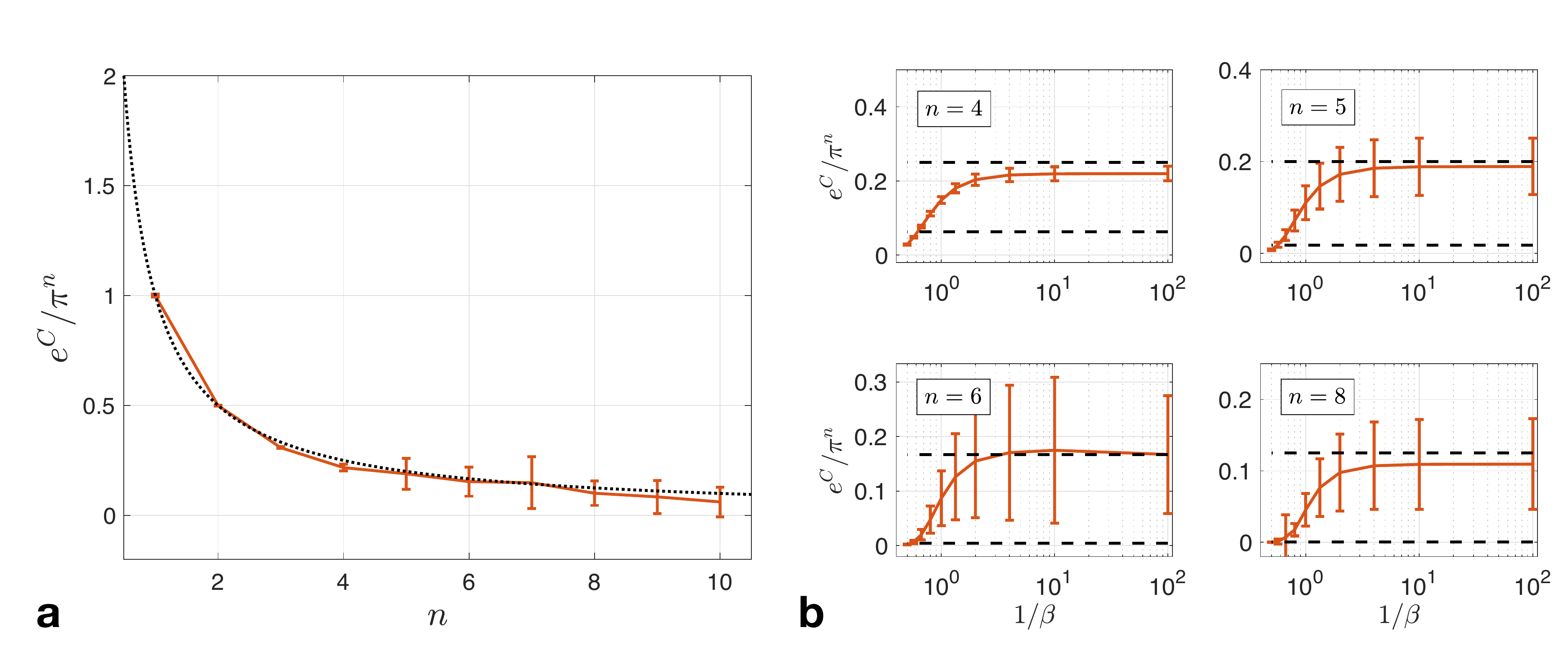}
\caption{\textbf{a)} Monte Carlo evaluations of the complexity assuming $\nu(\mathbf{x}) = 2^{-n}$, $\forall \mathbf{x}$ (red line). The error bars in the figures are standard deviations over $100$ independent estimates of the integral, each of which calculated using $10^5$ evaluation of the integrand. For comparison the function $1/n$ is plotted (black dotted line); \textbf{b)} The complexity at varying degrees of localisation of the input distribution from $\beta = 0.01$ (the inputs are observed in all possible configurations the same number of times) to $\beta = 2$ (the inputs are almost frozen around a specific configuration) for $n = 4,5,6$ and $8$. The complexity (red full line) increases, as expected, from $e^C/\pi^n = n^{-n/2}$ (bottom dashed line) to  $e^C/\pi^n = n^{-1}$ (top dashed line). 
Each estimate of the integral in the figure is the result of an average over 100 different Monte Carlo estimates, each of which has been calculated with $10^5$ points.}\label{fig:MC_result}
\end{figure}

The upper bound of the complexity, obtained by evaluating the integral with Monte Carlo simulations assuming a uniform distribution on the inputs, is shown in Figure \ref{fig:MC_result} a) at varying $n$, the number of inputs. Interestingly, the trend is consistent, within error bars, with the simple function $e^C = \pi^n/n$ (black dotted line) which has been also reported in Figure \ref{fig:MC_varyingLoc} (black dashed line) for comparison. As expected, the upper bound is found to be smaller than $\pi^n$.

Next, we assume that the input distribution follows an Ising model, namely $\nu(\bm{x}) \propto \exp(-\sum_{i<j}\beta J_{i,j}x_{i}x_{j} - \sum_i \beta h_i x_i)$, where $J_{i,j} = 1/n$ $\forall i \neq j$, $J_{i,i} = 0$ and $h_i = 0.1$, $\forall i$\footnote{The small parameter $h_i$ has been introduced to break the symmetry between the two configurations with all inputs equal.}. The localisation of the distribution is regulated by the parameter $\beta$: very small values of $\beta$ produce an almost uniform distribution whereas for larger values the distribution becomes more and more localised around the configuration with all inputs being one. As before, we used Monte Carlo simulations to estimate the integral for $n= 4,5,6$ and $8$ for different values of $\beta$. The results are shown in Figure \ref{fig:MC_result} b) with red lines. As expected, for small $\beta$ the curves approach the upper bound $e^C = \pi^n/n$ (top black dashed line) and when increasing $\beta$, the complexity decreases until some point around the limit $e^C = \pi^n/n^{n/2}$ (bottom black dashed line). 

As a final remark, in Appendix \ref{sec:field} we prove that, due to the parity of the hyperbolic cosine function, the complexity reaches the upper bound already when the input distribution is uniform over the configurations of a subset of $n-1$ inputs. This is evident in Figure \ref{fig:MC_varyingLoc} for the cases $n=2, 3$ and $4$: the complexity reached the upper bound already when all possible $2^{n-1}$ patterns relative to a subset of $n-1$ inputs were observed the same number of times. For larger values of $n$ this is not observed because none of the 10 different random paths in the simulation passes through the point in which the distribution is uniform over the $2^{n-1}$ patterns relative to a subset of $n-1$ inputs. In particular, as observed in Appendix \ref{sec:field}, this fact implies that the upper bound of the complexity for a model with $n$ inputs and without the bias term on the output variable is equal to that of a model with $n-1$ inputs and the bias.

\subsection{The emergence of statistical constraints in regularized models}\label{sec:asymptotics}

We now consider logistic models whose parameters $\theta_k$ are defined on a finite support, $| \theta_k | \le \Theta$, $\forall k$, with $\Theta$ being any finite positive number, and we study the effect of this effective regularization on the model complexity. In particular, we focus on the case of uniformly distributed inputs, $\nu(\bm{x}^\mu) = 2^{-n}$, $\forall \mu$, and compare the results obtained in the last section with un-regularized models. 

In Appendix \ref{sec:proof} we derive a fully analytical expression for the complexity of such models in the limit of an indefinitely large input layer. In this regime, the complexity can be approximated by the following equation
\begin{equation}\label{eqn:regularized}
e^C \sim \frac{2}{\sqrt{\pi n}}\left(\frac{8\pi\Theta^2 e^2}{n}\right)^{n/4}.
\end{equation}
As it is clear from the above expression, if the parametric family is defined in a box with half length $\Theta$, the volume occupied by the parametric family in the space of distribution will start shrinking after a certain critical dimension in contrast with the steady increase observed for un-regularized models, namely $e^C = \pi^n/n$ . More precisely, beyond the critical dimension, $n_c = 8\pi \Theta^2e^2$, increasing the number of inputs will make the model more constrained although still adding further degrees of freedom.

Intuitively this is due to the emergence of statistical constraints, namely the applicability of Central Limit arguments as shown in Appendix \ref{sec:proof}, which induces effective dependencies among parameters. In fact, in the proof of equation \ref{eqn:regularized} given in Appendix B, the assumptions of a large number of uniformly distributed inputs and parameters defined on a finite support allowed us to invoke Central Limit Theorem \footnote{Notice that the requirement of the finite support is the crucial assumption in the derivation and makes the difference with the previous cases of un-regularized models.} which implies that, for any choice of the parameter values, their linear combination with the inputs is constrained to follow a Gaussian distribution with zero mean. This is similar to what was observed in the previous section in the case of the most-localized input distribution where the complexity decreases with increasing the dimensionality of the model. In that case, the effective dependencies among parameters were induced by the correlations among inputs while, in this case, by statistical arguments.

Finally, the fact that the complexity starts decreasing after a certain critical dimension in regularized models is expected to be true for any non trivial input distribution given that in this section we proved it to be true for the uniform distribution which corresponds to the upper bound in the un-regularized case.

\section{Model selection}\label{sec:model_selection}
In this section, we are going to apply the outcomes of the previous sections to the problem of model selection with logistic regression models and binary variables. As outlined at the beginning of section \ref{sec:log_reg}, we are interested in solving the following problem: given $T$ measurements of the input and output variables in a logistic model with binary inputs, what is the model $\mathcal{M}$, i.e. the set of non-zero parameters, that has most likely generated the output data given the inputs? Since each weight parameter multiplies a different input in a logistic model, this is also an input selection problem: which subset of the inputs is relevant for predicting the output?
In the following, we will first derive a novel Model Selection criterion based on the posterior expansion and the results on the complexity obtained in section \ref{sec:log_reg}. Then we will apply this criterion to investigate the quality of model recovery and compare the performance with those achieved with other standard model selection recipes. As before and without any loss of generality, we will neglect the bias and consider only weight parameters, i.e. $b=0$ in equation \ref{eqn:logistic_regression} and $\bm\theta = \bm w$, since, if needed, the bias can always be introduced as a weight parameter multiplying an additional constant input.

\subsection{A novel model selection criteria}
Given $N$ potential predictors (inputs) for the output variable, there are $2^N$ possible models each of which trying to explain the output with a different subset of $n\leq N$ inputs. In other words, each model corresponds to a different subset of $n\leq N$ non-zero weight parameters. In a Bayesian Model Selection framework, we compare models according to their posterior probability. Assuming that all these models are a priori equally likely, this coincides with ranking models based on their likelihood $p(\hat{y}|\hat{\bm{x}},\mathcal{M})$. In section \ref{sec:log_reg} we saw that $p(\hat{y}|\hat{\bm{x}},\mathcal{M})$ can be approximated as $p(\hat{y}|\hat{\bm{x}},\mathcal{M}) = T\ell(\bm{\theta^\ast}) + \log r + O(1/T)$ where the first term is the likelihood of the data calculated at the Maximum Likelihood estimates of the parameters and the second term is a penalization factor. The latter is given by $\log r =  -\frac{n}{2}\log\left(\frac{T}{2\pi}\right) - C $ and consists of the sum of the BIC and the complexity term respectively.  In the last section we found that the value of $C$ depends on both the localisation of the input distribution and the dimensionality of the model; it is upper bounded by $\log(\pi^n/n)$; finally, it decreases down to some value around $\log(\pi^n/n^{n/2})$ as the localisation of the input distribution increases. In the first case, the penalisation would be $\log r = -\frac{n}{2}\log\left(\frac{\pi T}{2}\right) + \log n$ which is dominated by a linear dependence on n and a logarithmic dependence on T, similar to the case of BIC. In the second case, the penalisation would be $\log r = -\frac{n}{2}\log\left(\frac{\pi T}{2 n}\right)$ which is a weaker penalisation with respect to the previous case.

We now introduce the entropy $H_{n}(\bm x) = -\sum_{\bm x} \nu (\bm x) \log_2 \nu (\bm x)$ as a measure of the localisation of the input distribution: the value of the entropy approaches $n$ (the dimension of the model) when the distribution of the inputs approaches the uniform distribution; whereas it shrinks to some value around $\log_2 n$ when the distribution is localised on approximatively $n$ linearly independent configurations.

Based on the results of the previous section and on the aforementioned measure of localisation, we introduce the following heuristic for calculating the penalization of a model with $n$ inputs whose distribution has entropy $H_{n}$:
\begin{equation}\label{eqn:novel_model_sel_criterion}
\log r =  -\frac{n}{2} -\frac{n}{2}\log\left( \frac{ T H_{n}}{n H_{N}} \right) + \log n,
\end{equation}
where $H_N$ is the entropy of the full design matrix with all $N$ predictors taken into account. We further motivate this formula in Supplemental Information (section 2), while, in the following, we provide the intuition behind it and discuss its properties and performance on model selection tasks.

Equation \ref{eqn:novel_model_sel_criterion} allows us to approximately interpolate between the expected trends as derived in the limiting cases of a uniform and localised distribution. In fact, when $H_n \simeq n$, the largest contribution to the penalization exhibits the expected BIC-like trend, $\log r \sim -\frac{n}{2}\log\left(T\right)$, which represents the leading contribution in this regime given that in order to have $H_n \simeq n$, a sample size larger than the dimensionality of the configuration space is needed, i.e. $T \geq 2^n$, and given that $H_N \leq \log_2 T$. Alternatively, one might think about this as the type of penalization which affects mostly models with $n \ll \log_2 T$ (or sparsity $s = 1-n/N \gg 1-  \log_2 T/N$).

On the other hand, when increasing the localisation of the distribution, the entropy growth with $n$ would be drastically reduced. This is typically the case for models with a large number of inputs, namely with $\log_2 T \ll n \leq N$, where $H_n \simeq H_N$ (see also Figure 3 in Supplemental Information). In this case, the factor $T$ in the argument of the logarithm in equation \ref{eqn:novel_model_sel_criterion} is counterbalanced by a large value of $n$ leading to a weaker penalization, similarly to what was obtained when substituting the lower bound of the complexity into the posterior expansion. Moreover, for very dense models $n \approx N$ and small datasets $T \approx N$, the leading term of the penalization tends to an AIC-like one of order $O(n)$.

\subsection{Model selection tests}
In general, we expect that results delivered by different model selection recipes will differ more among each other as the dimensionality of the models grows large. For example, the difference between the AIC and BIC penalization factors grows as $O(n\log T)$. This is an important issue for real world datasets for which $T = k \cdot N$ with $N\gg 1$ and $k \geq 1$. In our tests, we consider $N = 50$ binary inputs ($N=100$ in Supplemental Information) and $k = 5, 50$ and $200$. As before, in section \ref{sec:mathematics}, we generate samples for the inputs according to an Ising model, $p(\bm{x}) \propto \exp(-\sum_{i<j}\beta J_{i,j}x_{i}x_{j} - \sum_i \beta h_i x_i)$, where $J_{i,j} = 1/N$, $\forall i \neq j$, $J_{i,i} = 0$ and $h_i = 0.1$, $\forall i$, where the small parameters $h_i$ have been introduced to break the symmetry between the two configurations with all inputs being equal. The choice of an Ising distribution is motivated by the fact that it is easy to tune the localisation of the distribution by varying the parameter $\beta$. 

Given the input data, we generate the output according to the conditional probability distribution $p(y|\bm x)$ of equation \ref{eqn:logistic_regression} employing only a random subset $n\leq N$ of the predictors. This is achieved by assigning non zero values in the parameter vector only to the weights multiplying the selected predictors. The number of selected predictors, $n$, is decided according to the desired level of sparsity, $s = 1-n/N$. Each non zero weight is drawn independently from the distribution defined as: $P(w) = 1/2$, for $|w-1|\leq1/2$ and $|w+1|\leq1/2$; $P(w) = 0$ otherwise. After that, the parameter vector is normalized by $\sqrt{n}$ to ensure that the effective field acting on the output is of $O(1)$ regardless of the number of non-zero parameters in the model.

\begin{figure}
\centering
\includegraphics[width = \columnwidth]{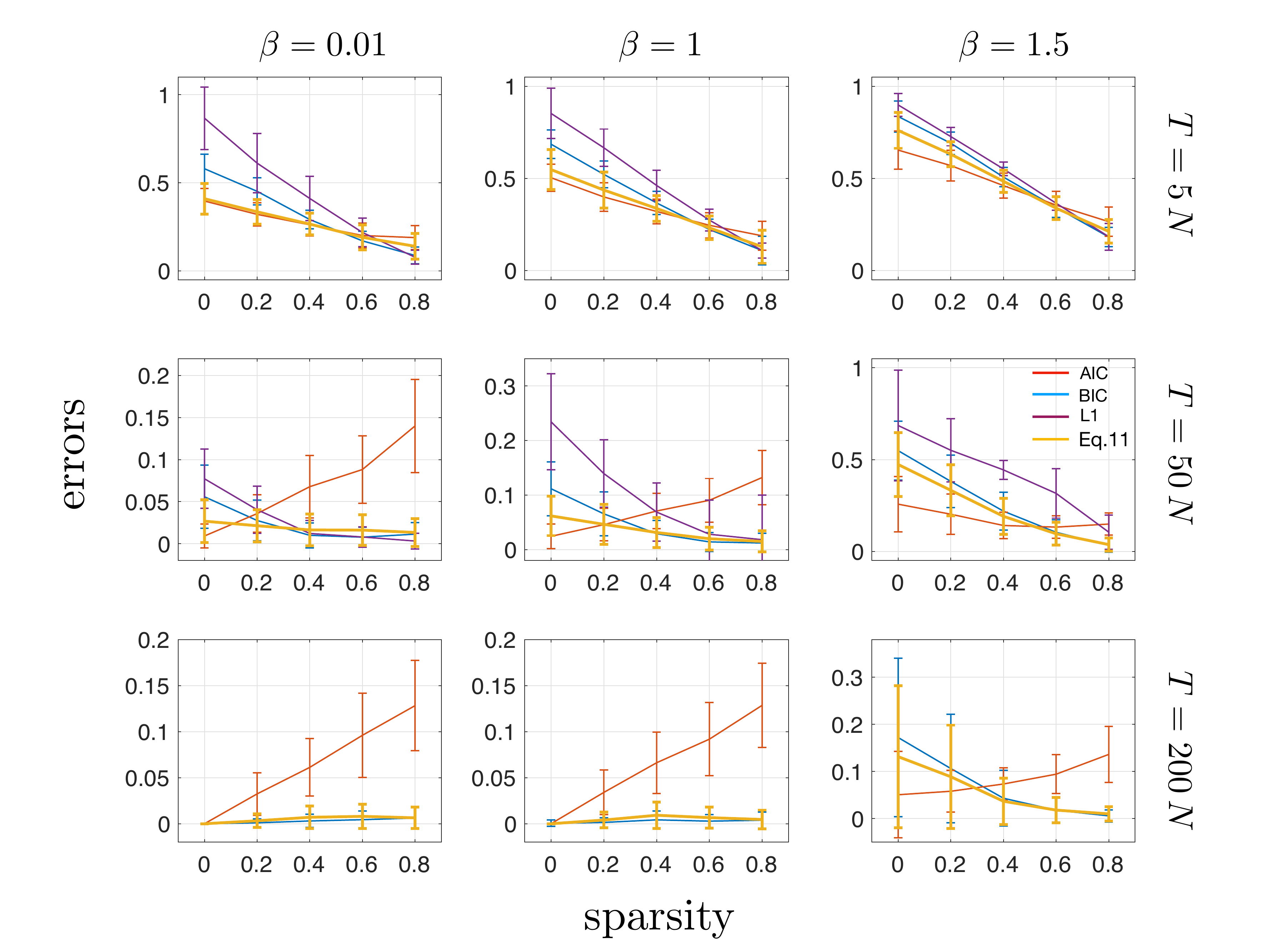}
\caption{The reconstruction errors versus the sparsity of the ground truth and for different levels of the localisation of input distribution $\beta$ (columns) and sample size $T$ (rows). The errors are the mean fraction of misclassified active/inactive parameters over $100$ independent realisations of the same experiments and the error bars are the corresponding standard deviations. The size of the input layer is $N = 50$. We compared performance employing different model selection criteria: AIC (red), BIC (blue), equation \ref{eqn:novel_model_sel_criterion} (yellow) and logistic regression with $\ell_1$ regularization (purple) (the latter only for $T = 5 N$ and $T=50 N$).}\label{fig:model_selection}
\end{figure}

Thus, after having defined the ground truth and generated $T$ samples from it, we aim at recovering its structure from the data by doing model comparison: we rank the models according to their maximum likelihood value and penalise them for their complexity by employing equation \ref{eqn:novel_model_sel_criterion}. We compared the performance with those obtained with some of the most common approaches, i.e BIC \citep{BIC}, AIC \citep{Akaike}, $\ell_1$ regularization \citep{Lasso, Kim07} with a {\itshape K-fold} cross validation procedure for selecting the regularizer ($K = 5$).
Except for the $\ell_1$ method where the search in the space of models is embodied in the optimization, for the other methods we need to define the pool of candidate models. In fact, the set of all possible candidates grows exponentially with the number of predictors so that an extensive search among all possibilities is impracticable for large values of $N$. We therefore resort to approximate procedures to walk through this huge model space. We use a decimation procedure \citep{lecun1990optimal,hassibi1993second,Decelle13}: we start from a full model with all predictors $N$ connected to the output and estimate its parameters; then the next model to include in the pool would employ all predictors as in the previous model except for the one whose corresponding interaction parameter had been inferred as having the least absolute value; at each step, we define new models by removing the parameter with the smallest absolute value until we reach the model with no active interactions at all, namely the model with the output independent from all candidate predictors. We have also tried a ``forward'' approach by starting from the null model and adding at each step the parameter that would cause the biggest jump in the likelihood. The reconstruction errors with both procedures are found to be almost indistinguishable in our simulations and, therefore, we report only the results with the decimation procedure. These errors are plotted in Figure \ref{fig:model_selection} versus the sparsity of the ground truth and for different levels of the localisation of the input distribution $\beta$ (columns) and sample size $T$ (rows). The errors are the mean fraction of misclassified non-zero/zero parameters over $100$ independent realisations of the same experiments and the error bars are the corresponding standard deviations. In the figure, the sparsity level varies from $0$ to $0.8$ meaning that the ground truth in the first case is a model with all weights being non-zero, whereas in the second case $80\%$ of them are set to zero.

Clearly, the reconstruction error decreases as the sample size $T$ increases for all methods, but more slowly for AIC. On the other hand, at increasing $\beta$, model retrieval becomes harder. Consistently, the worst reconstruction is observed for the largest value of $\beta$ and the smallest sample size ($\beta = 1.5$ and $T = 5 N$ in Figure \ref{fig:model_selection}). In the latter case, the likelihood is quite flat so that a very sparse model is typically selected by any model selection criterion. In fact, the information on the activity of the output provided by a new predictor which is highly correlated with the ones already employed in the model is expected to be small because much of its variability is already encoded in the set of predictors in use. This is true even if this predictor is indeed present in the ground truth. On the other hand, when all the predictors are maximally different from each other (maximal information conveyed by each predictor) and the data length is large, the likelihood is expected to be quite sharp around the ground truth, allowing for a very good reconstruction. This is indeed the case of $\beta = 0.01$ and $T = 200 N$ in Figure \ref{fig:model_selection} where each model selection recipe achieved its minimal reconstruction error.

BIC and $\ell_1$ regularization employ a stronger penalization than AIC which explains why in the figure the latter methods are always better than AIC in recovering sparse models, but worse when it comes to unveiling dense models. The proposed penalization in equation \ref{eqn:novel_model_sel_criterion} acts against this unbalance in the recovery errors between sparse and dense models. In fact, it tends to match the BIC trend for high sparsity and the AIC one for low sparsity, resulting in the best model selection recipe in our simulations: in Figure \ref{fig:model_selection}, for any level of sparsity, the reconstruction error with our method is the smallest or gets close to the smallest one achieved by the other methods whose performances instead strongly depend on the sparsity of the ground truth. In Supplemental Information (Figure 4), numerical experiments with $N=100$ further substantiates these results. This result is very important in applications given that normally we are not given any information on the expected number of predictors (sparsity) and therefore we might want to treat sparse and dense models on the same footing. As shown in this section, our method achieves this goal whereas standard alternatives typically introduce a bias toward some sparsity range. 

\section{Degenerate models}\label{sec:degenerate_models}
In this section we consider {\itshape degenerate models}, namely those models for which some components of the parameter vector ${\bm\theta}$ are constrained to be equal. Evaluating the complexity for degenerate models is particularly interesting for model selection given that usually model selection criteria, such as Bayesian Information Criteria and Akaike Information Criteria, can distinguish models only based on the number of parameters employed: for instance, a model with one parameter connecting one input to the output would be equivalent to a model connecting $n$ inputs through the same parameter.
For logistic regression models, it is easier to see that any degenerate model, i.e. any model built from a non-degenerate model by explicitly constraining a subset of parameters to be equal, can be mapped into a non-degenerate model with categorical inputs \footnote{From equation \ref{eqn:logistic_regression}, the predictor of a degenerate parameter is the sum of all binary predictors connected to the output through the same parameter, i.e. a categorical predictor.}. Therefore the additional complexity of a degenerate model with respect to a non degenerate one with the same number of parameters is related to the additional complexity conveyed by categorical inputs with respect to binary ones.

In the following we will consider a simple example of such models and discuss an application on a real dataset.
\subsection{1-parameter degenerate models}
The simplest example of such models is a logistic regression model with $n$ binary inputs $x_i$ with $i = 1,..., n$ and the corresponding $n$ weights tied together through the same parameter $\theta$:
\begin{equation}\label{eqn:degenerate}
p(y|\bm{x}) = \frac{e^{y \theta (\sum_{i=1}^n x_i - b)}}{2\cosh(\theta(\sum_{i=1}^n x_i - b))}
\end{equation}
where $b$ is a fixed parameter which works as a threshold.
Given $T$ observations of the inputs and the output $X^{(t)} = \{y^{(t)}, \bm{x}^{(t)}\}$, for $t = 1,..., T$, the normalised log-likelihood of the model is given by 
\begin{equation}\label{eqn:likelihood_degenerate}
\ell (\theta) = \theta\overline{y (X_n-b)} - \overline{\log(2\cosh(\theta (X_n-b)))},
\end{equation}
where $X_n = \sum_{i=1}^n x_i$. As the form of the likelihood suggests, this model can be mapped into a model with only one categorical input $X_n \in \{-n,-n+2, ..., n-2, n\}$, connecting to a binary output $y$ through the parameter $\theta$. Consequently the Fisher Information takes the form $F(\theta) = \sum_{X_n} \nu(X_n) \,{(X_n-b)}^2\cosh^{-2}(\theta (X_n-b))$, where $\nu(X_n)$ represents the distribution of the sum of random variables $x_i$, and the complexity is simply the integral of its square root. As we did in section \ref{sec:mathematics} for non-degenerate models, we resort to triangular inequality to study the bounds of the complexity by varying the distribution $\nu(X_n)$ and the dimension $n$. This procedure leads to the simple inequality:
\begin{equation}\label{eqn:upper_degenerate}
e^C \leq \pi \sum_{X_n} \sqrt{\nu(X_n)}.
\end{equation}
It can be easily shown that the upper bound in the right hand side of the previous equation is maximised when the distribution is uniform over all possible outcomes of the variable $X_n$ which means $\nu(X_n) = 1/(n+1)$. By substituting this form of the distribution in equation \ref{eqn:upper_degenerate}, we find that $e^C \leq \pi \sqrt{n+1}$. Thus the complexity grows at most as the square root of the number of inputs which is a much slower growth compared with the exponential one in non-degenerate models. Therefore, it is clear that the additional complexity gained by growing the alphabet of a categorical variable is generally very small compared to the additional complexity obtained by adding more independent parameters. 

On the other hand, the minimum value for the right hand side of equation \ref{eqn:upper_degenerate} will be reached when only one particular value, say $\bar{X}$, of the variable $X_n$ is observed, namely when $\nu(X_n) = \delta_{X_n,\bar{X}}$, where $\delta$ stands for the Kronecker delta function. In this case, the resulting complexity is exactly $e^C = \pi$, as we saw for a one-parameter model with a binary input in section \ref{sec:case_n=2}.

An interesting case is when all inputs are uniformly distributed so that in a large sample ($T\gg 2^n$) we expect to observe all possible $2^n$ configurations the same number of times. In this case, the number of occurrences of a particular value of $X_n$ is given by $\binom{n}{(n+X_n)/2}$ \footnote{The number of plus ones in a pattern with $X_n = x$ is $(n+x)/2$; it follows that the combinatorial factor $\binom{n}{(n+x)/2}$ counts all possible patterns with the same number of plus ones, i.e. the same value of $X_n$.}. It follows that the distribution of $X_n$ will be $\nu(X_n) =\binom{n}{(n+X_n)/2}2^{-n}$. By plugging this distribution into equation \ref{eqn:upper_degenerate}, we find $e^C \leq \pi \, 2^{-n/2} \sum_{k=0}^n \sqrt{\binom{n}{k}}$. The right hand side of the inequality is very well approximated by the function $\pi (a_1\cdot n + a_2)^{1/4}$ with $a_1 = 2\pi$ and $a_2 = \pi/\sqrt{2}$ which gives a better intuition of the behaviour of the complexity in this particular case: it increases with $n$ but much slower than the upper bound; it is always above the lower bound identified by $\pi$ and, as $n$ goes to zero, it tends to the same value taken by the upper bound. Notice that both the upper bound and this particular solution do not tend to $e^C=\pi$ for $n=1$ because these results are obtained by employing triangular inequality.

\begin{figure}[t!]
\centering
\includegraphics[scale = 0.5]{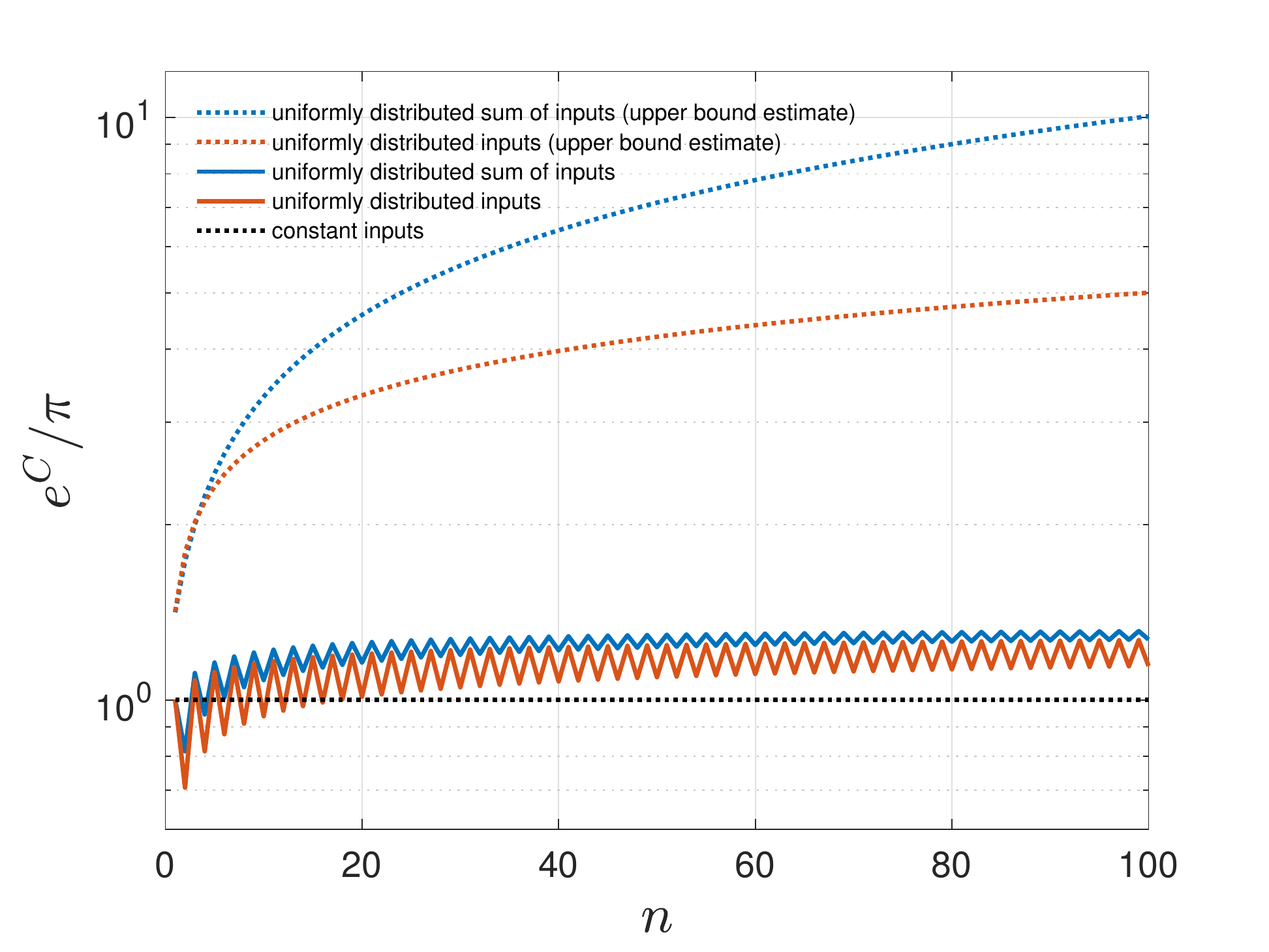}
\caption{The complexity of 1-parameter degenerate models versus the degree of degeneracy $n$ of the parameter. The complexity has been evaluated numerically using equation \ref{eqn:degenerate} and $b=0$ for two cases: when all possible values of the sum of inputs are observed the same number of times (solid blue line) and in the case in which all possible input configurations are instead observed an equal number of times (solid red line). We also report the upper bounds estimate in the aforementioned two cases (dotted lines, same color coding) and the value of the complexity when the distribution is extremely localized (black dotted line), i.e. when the input layer is frozen in a certain configuration.}\label{fig:complexitydeg}
\end{figure}

We present these analytical results in Figure \ref{fig:complexitydeg} (dotted lines) at varying the size of the input layer $n$. In the same figure, we compare these trends with simulations obtained by evaluating the one dimensional integral $e^C = \int d\theta \sqrt{F(\theta)}$ numerically (solid lines). In evaluating the integral we fixed $b=0$ and studied it at varying $n$. As it is clear from the figure, the complexity oscillates while very slowly increasing with $n$ in both cases of uniformly distributed random inputs (red line) and uniformly distributed sum of inputs (blue line). The oscillation is due to the fact that when $n$ is even, the categorical variable $X_n$ can take exactly the value of the threshold $b$, which has been set to zero. The complexity increases with the dimension $n$ more rapidly for models with a small number of predictors, whereas its value becomes almost independent of $n$ for large models. Moreover, it is interesting to notice that the complexity of a one parameter degenerate model for $n<100$, as shown in \ref{fig:complexitydeg} (dotted line), is always smaller than the maximum value of the complexity for a two parameter non degenerate model which is $e^C = \pi^2/2$, as we have seen in section \ref{sec:case_n=2}.

As a conclusion, given that a one parameter degenerate model with $n$ inputs can be mapped into a model with only one categorical input with an alphabet of $n+1$ symbols, the additional complexity conveyed by categorical inputs with respect to binary ones increases very slowly with the cardinality of the alphabet of the categorical variable, showing a larger effect for small values of $n$, and it was found to be lower than the increment in complexity which would follow from adding another non degenerate parameter to the model, even for large values of $n$.

\subsection{The 13 keys to the White House}
An interesting application is the question of predicting the U.S. presidential elections from the values of some predictors. A successful method for forecasting the verdict of US presidential elections is ``The 13 keys to the White House'' \citep{Lichtman81}. The method was developed in the eighties and, since then, has been able to predict correctly presidential elections \citep{Lichtman2016}.  It is based on 13 binary questions (keys) and each time the answer to a key is false, the key is turned against the party in power. If six or more keys are false, the challenging party is predicted to win the elections otherwise the party holding the White House will be more likely to win. Since the predictions depend only on the sum of the keys, i.e. how many of them are false, the method can be interpreted as the deterministic counterpart of a 1-parameter degenerate model (see Supplemental Information, section 3).

The question we would like to ask from a Bayesian Model Selection point of view is: is there any ``simpler" explanation of the outcomes of presidential elections which would perhaps involve less predictors? We compare all possible models corresponding to all possible ways of selecting a subset of predictors from the original set of the 13 keys and use that subset for forecasting the presidential elections (see Supplemental Information). This amounts at comparing $2^{13}$ models. Since all models, except the model employing no predictors at all, can be cast into degenerate models with only 1 parameter, they will all have the same BIC penalization factor. It follows that the largest penalization factor which differs among models is represented by the complexity.

\begin{figure}[t]
\centering
\includegraphics[width= \columnwidth]{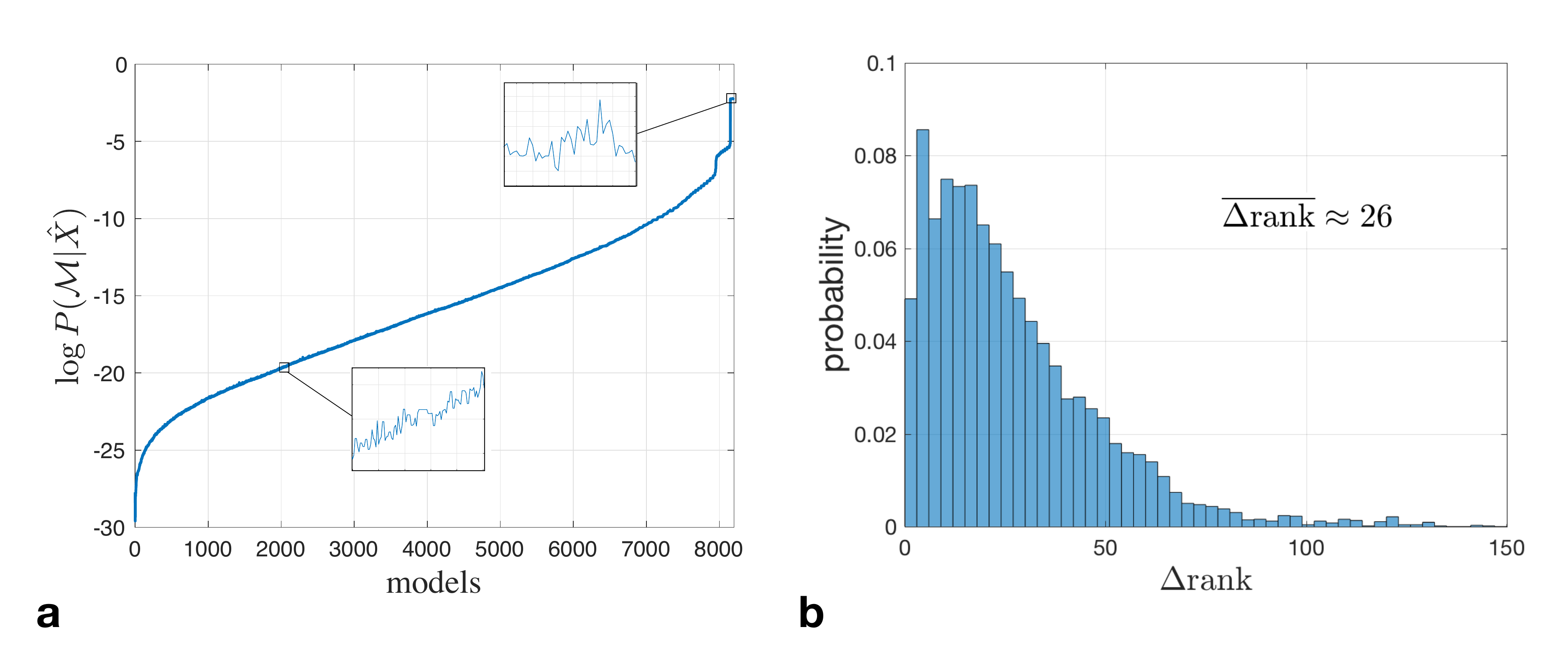}
\caption{\textbf{a)} The values of the posterior for all the models ranked according to their maximum likelihood; \textbf{b)} Distribution of $\Delta rank$ which is the order of magnitude of local shuffling induced by the complexity on the likelihood based ranking of models.  
}\label{fig:degenerate_results1}
\end{figure}

In Figure \ref{fig:degenerate_results1} a) we plot the posterior probability $p({\cal M}|\hat{X})$ for all models ranked according to their maximum likelihood values. The figure clearly shows that the effect of taking into account the complexity on model selection is local, i.e. it concerns models which are very close in likelihood. In fact, the posterior seems to be monotonically increasing on a larger scale while it presents fluctuations on a finer scale (insets). 
Therefore, the complexity induces a local reordering of the models with respect to the likelihood ranking. We measure the range of such an effect by introducing the quantity $\Delta rank$. Specifically, given all models ranked in a non-decreasing order according to their likelihood, $\Delta rank$ is the range of models surrounding a given one within which the fluctuations of the complexity are approximately equal to the variation of the likelihood.
In some sense $\Delta rank$ is a measure of the uncertainty induced by the complexity in the likelihood ranking. It varies across models with a mean value of $\overline{\Delta rank} \approx 26$, as shown in Figure \ref{fig:degenerate_results1} b).

\begin{figure}[t]
\centering
\includegraphics[width= \columnwidth]{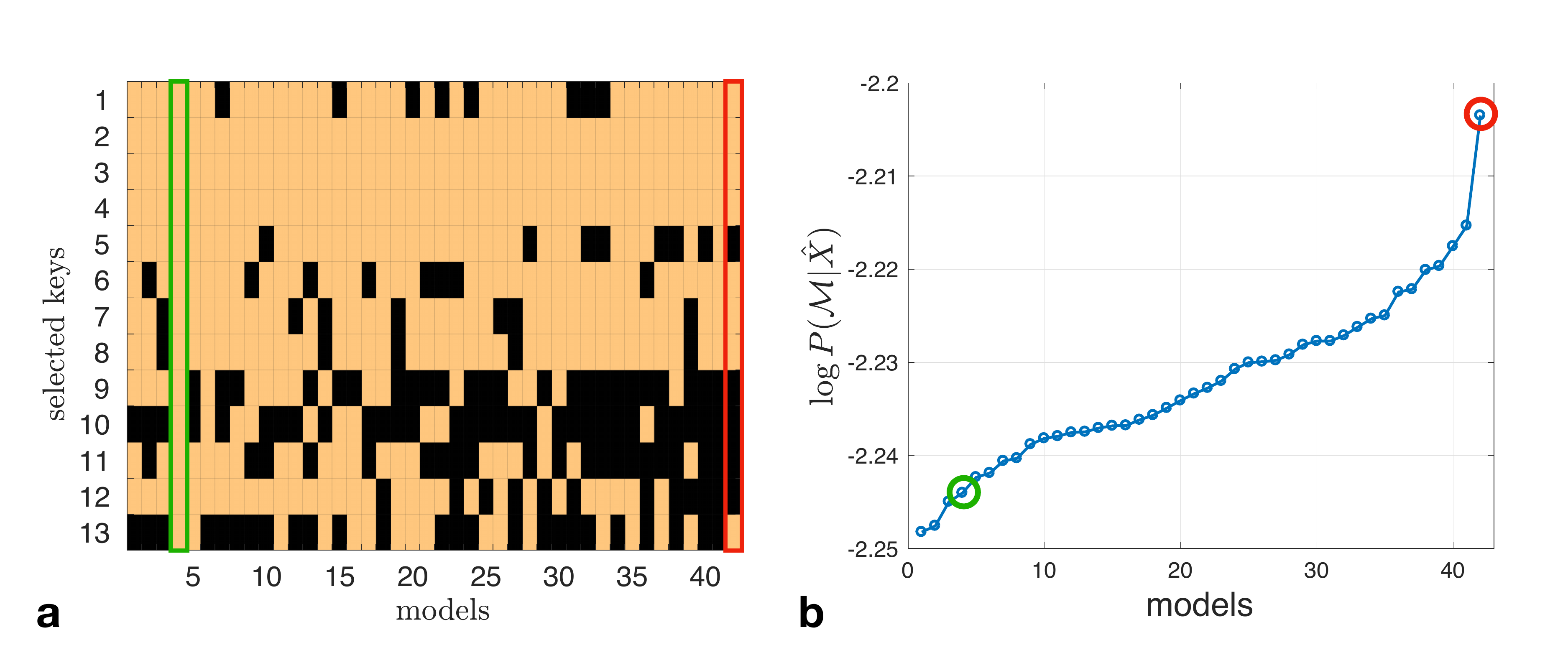}
\caption{\textbf{a)} The 42 models with the largest posterior. The models are ranked by increasing values of the posterior. Each model is represented by a column vector with binary values: yellow if the key is active in the selected model and black otherwise. The model of interest are highlighted: model 4 (green) which correspond to the model with all 13 keys and model 42 (red) which is the one that maximize the posterior; \textbf{b)} The posterior values for the 42 most likely models.}\label{fig:degenerate_results2}
\end{figure}

Moreover, Figure \ref{fig:degenerate_results1} a) reveals that, interestingly, there are some models in our pool that are attached a posterior significatively larger than the others. There are 42 of such models and they are reported in Figure \ref{fig:degenerate_results2} a). These models fit the dataset exactly ($\ell \approx 0$). It follows that, given that the BIC term is the same overall, the complexity is the only term varying across models. In Figure \ref{fig:degenerate_results2} b) the models are ranked according to their posterior (or equivalently from the most to the least complex model, given that the likelihood and the BIC term are constant). As mentioned in the previous section, the complexity does not depend merely on the number of predictors but also slightly on the distribution of the corresponding categorical variable attached to them. In fact, the model with all 13 keys active turns out to be not the most complex model although still among the most complex ones. On the other hand, the model with the largest posterior overall, contains only 8 out of the 13 keys (Keys  1,2,3,4,6,7,8,13) where the first seven keys are among the most frequent keys over all 42 models. Interestingly, by looking at the set of predictors not included in the best model (Keys 5,9,10,11,12), we could see that: according to this model, it does not matter for the forecast of the next presidential election whether the economy is in recession during the electoral campaign (key 5), but what does matter is rather the economic growth during the whole previous term (key 6); military failures or successes are irrelevant as predictors (keys 10 and 11); finally, it does not matter whether the incumbent administration is untainted by a major scandal (key 9) or whether the incumbent-party candidate is charismatic or a national hero while it is relevant whether the challenging-party candidate is charismatic (key 13)\footnote{We refer to \citep{Lichtman2016} for a detailed explanation of the meaning of the keys.}
 
As a conclusion, by comparing all possible degenerate models with different predictors through a Bayesian model selection approach, we found that the original method of the 13 keys to the White House is not attached a posterior probability larger than many other candidates. Instead, a model with only 8 predictors is the one that achieves the largest posterior. However it is worth noting that the differences in the values of the posterior among these 42 most likely models are very tiny and eventually the result of the model comparison would strongly depend on the choice of the prior over models. Our choice of the prior favours the least complex models among the best candidates, i.e. the one which account for the minimum number of distinguishable probability distributions.

\section{Conclusions}

In this paper we address the question of estimating the complexity of a logistic regression model with binary inputs. As shown in \cite{balasubramanian1997statistical}, the complexity of a model can be related to the number of distinguishable distributions that a model can represent. In this perspective, it is defined as the integral of the square root of the determinant of the Fisher Information matrix. We evaluate analytically the integral in special cases and corroborate the analysis with Monte Carlo simulations. We find that the integral depends on the degree of correlation of the inputs: correlations among inputs induce effective dependencies on the parameters thus reducing the volume that the model occupies in the space of distributions. It follows that when the inputs are maximally uncorrelated, the model achieves its maximal value of the complexity, whereas when the inputs are highly correlated, it becomes highly constrained and it is characterized by a low value of the complexity. The upper bound of the complexity has been estimated with Monte Carlo methods. We find that it can be very well approximated by the simple relation $e^C = \pi^n/n$, where $n$ is the number of parameters. The lower bound instead is reached when the number of unique patterns observed in the input layer is equal to the number of parameters. In the latter case, we find an exact solution for the value of the complexity, $e^C = \pi^n \prod_{i =1}^n \sqrt{\nu (\bm{x}^i)}$. The solution depends on the distribution $\nu (\bm{x}^i)$ of these $n$ patterns and ranges from $e^C = \pi^n \sqrt{(T - n + 1)/T^n}$ to $e^C = \pi^n/n^{n/2}$.  Monte Carlo estimates of the integral at varying degrees of correlation (localisation of input distribution) further confirm these results. Finally when the number of unique patterns observed in the input layer is less than the number of parameters, the data matrix is not full rank and $e^C = 0$: the model is constrained into a lower dimensional manifold in the space of distribution induced by the deterministic relation among parameters.

Moreover, we investigate the complexity of models whose parameters are defined on a finite support (regularized models). In the case of maximally uncorrelated inputs, we find that the complexity starts decreasing with the dimensionality of the model after a certain critical dimension as opposed to the steady exponential increase, i.e. $e^C = \pi^n/n$, found for models with parameters defined on the entire real axis. This is due to the gradual appearance of dependencies among parameters induced by emergent statistical constraints. Thus, for large values of $n$, increasing the number of parameters in regularized models will always increase the degrees of freedom of the model, but it will also make the model more constrained (less complex), at odds with the corresponding un-regularized case.

We apply these results on the model complexity for devising a novel model selection criterion that would take into account the correlations among inputs when deciding how much a model should be penalised in a Bayesian model selection framework. To this purpose, we introduce a heuristic for the penalization cost based on the entropy of the input distribution. Consistently, our proposal assigns a strong penalization, similar to a BIC-like one, to models with a large value of the complexity and a weaker one, more similar to an AIC-like penalization, to extremely constrained models. We then tested this ansatz by simulating data from logistic models with various levels of sparsity, sample size and input correlations, and then asking different model selection criteria to retrieve the correct model. Numerical tests clearly show that our proposal always achieves (or gets very close to) the smallest reconstruction error for all levels of sparsity, data size and input correlations whereas performances with the other tested methods strongly depend on the sparsity of the ground truth. As a consequence, our method qualifies as the best choice in our tests and would be especially useful in those cases in which a priori we have no information about the sparsity of the ground truth. As an example, this might be the case for a population of neurons simultaneously recorded from some region of the brain. In fact, the sparsity of the network of dependencies among neurons strongly varies according to if the recorded cells belong to a functionally connected module or many independent ones as well as according to how the behavioural task during the recording shapes the activity of the cells \citep{Dunn2015}. These information might often be not known to the scientist a priori.

Finally, we address the question of how the alphabet size of the inputs affect the value of model complexity. The problem is equivalent of studying the complexity of a model with binary inputs, but with degenerate parameters, namely when subsets of parameters are forced to be equal. In this respect, we study analytically and numerically the simple case of a model with only one parameter by varying its degree of degeneracy or, alternatively, the alphabet size of the corresponding effective input. We find that the additional complexity conveyed by categorical inputs with respect to binary ones increases very slowly with the alphabet size of the categorical variable, showing a slightly larger rise for small sizes, and it is always lower than the increment in complexity which would follow from adding another non degenerate parameter to the model.

Studying the complexity of degenerate models is also interesting because many commonly used model selection recipes are based on penalising models according to some cost function which often depends only on the number of parameters and sample size. A Bayesian Model selection approach would instead employ a penalization which might differ even for models with the same number of parameters. As an application, we study the dataset of the 13 keys to the White House \citep{Lichtman81,Lichtman2016}. In fact, the question of deciding if a model with less predictors would be better at explaining the variability in the dataset with respect to the full model with all 13 predictors, can actually be cast into a model selection problem with 1-parameter degenerate models. As one would expect, the contribution of the complexity is very small as compared with that coming from the likelihood. Yet, we show that it becomes relevant for deciding among the subset of models with the largest value for the likelihood. 

It would be interesting to extent this study of the complexity to a more general class of models known as Generalized Linear Models \citep{Nelder1972} which is widely employed in applications, for instance in neuroscience for studying the neural code of cells or populations of cells in the brain \citep{roudi2015multi}. 

Another interesting application is related to the fact that one of the most powerful approaches for inference and structure learning in Ising model is based on the so-called pseudo-likelihood approximation \citep{ravikuumar10, Aurell12, Decelle13}. This approximation basically consists of substituting the exact likelihood function, whose optimization is hard because of the partition function, with a sum of conditionally independent logistic regression functions with binary inputs and outputs. Within this approximation, based on the results of section \ref{sec:model_selection}, we would expect that our proposed criterion (equation \ref{eqn:novel_model_sel_criterion}) will deliver close to optimal performances for nodes with any level of degree distribution (sparsity) and, therefore, on a wide range of network topologies, including topologies where highly connected nodes coexist with almost isolated ones, as for instance in scale free and small world networks. 

Finally, it must be noticed that equation \ref{eqn:novel_model_sel_criterion} is a heuristic inspired by the truncated posterior expansion of equation \ref{eqn:bayesian_approximation} so it should inherit the limitations buried in the truncation. In particular, terms of order 1/T, which are neglected in equation \ref{eqn:bayesian_approximation}, might be relevant in some regimes \citep{balasubramanian1997statistical}. As shown also by our numerical tests, when the sample size approaches the dimensionality of the model or for large input correlations, model selection represents a hard task and performances are generally poor with any approach. It would be interesting to study higher order terms in the expansion and see whether the information contained in them could be used to improve model selection in this regime.

\section{Acknowledgments}
The authors are grateful to Prof. Benjamin Dunn for reading through the paper.

\medskip

\bibliographystyle{apalike}

\begin{thebibliography}{}

\bibitem[Akaike, 1974]{Akaike}
Akaike, H. (1974).
\newblock A new look at the statistical model identification.
\newblock {\em Automatic Control, IEEE Transactions on}, 19(6):716--723.

\bibitem[Aurell and Ekeberg, 2012]{Aurell12}
Aurell, E. and Ekeberg, M. (2012).
\newblock Inverse ising inference using all the data.
\newblock {\em Phys. Rev. Lett.}, 108:090201.

\bibitem[Balasubramanian, 1997]{balasubramanian1997statistical}
Balasubramanian, V. (1997).
\newblock Statistical inference, occam's razor, and statistical mechanics on
  the space of probability distributions.
\newblock {\em Neural computation}, 9(2):349--368.

\bibitem[Battistin et~al., 2015]{Battistin2015}
Battistin, C., Hertz, J., Tyrcha, J., and Roudi, Y. (2015).
\newblock Belief propagation and replicas for inference and learning in a
  kinetic ising model with hidden spins.
\newblock {\em Journal of Statistical Mechanics: Theory and Experiment},
  2015(5):P05021.

\bibitem[Besag, 1972]{Besag1972}
Besag, J.~E. (1972).
\newblock {Nearest-Neighbour} systems and the {Auto-Logistic} model for binary
  data.

\bibitem[Billingsley, 1995]{Billingsley95}
Billingsley, P. (1995).
\newblock {\em {Probability and Measure}}.
\newblock Wiley, 3 edition.

\bibitem[Bulso et~al., 2016]{Bulso2016}
Bulso, N., Marsili, M., and Roudi, Y. (2016).
\newblock Sparse model selection in the highly under-sampled regime.
\newblock {\em Journal of Statistical Mechanics: Theory and Experiment},
  2016(9):093404.

\bibitem[Chen et~al., 2008]{chen2008}
Chen, M.-H., Ibrahim, J.~G., and Kim, S. (2008).
\newblock Properties and implementation of jeffreys’s prior in binomial
  regression models.
\newblock {\em Journal of the American Statistical Association},
  103(484):1659--1664.

\bibitem[Cox, 1958]{cox1958regression}
Cox, D.~R. (1958).
\newblock The regression analysis of binary sequences.
\newblock {\em Journal of the Royal Statistical Society. Series B
  (Methodological)}, pages 215--242.

\bibitem[Decelle and Ricci-Tersenghi, 2014]{Decelle13}
Decelle, A. and Ricci-Tersenghi, F. (2014).
\newblock Pseudolikelihood decimation algorithm improving the inference of the
  interaction network in a general class of ising models.
\newblock {\em Phys. Rev. Lett.}, 112:070603.

\bibitem[Dunn et~al., 2015]{Dunn2015}
Dunn, B., M{\o}rreaunet, M., and Roudi, Y. (2015).
\newblock Correlations and functional connections in a population of grid
  cells.
\newblock {\em PLoS computational biology}, 11(2):e1004052.

\bibitem[Friedman et~al., 2001]{friedman2001elements}
Friedman, J., Hastie, T., and Tibshirani, R. (2001).
\newblock {\em The elements of statistical learning}, volume~1.
\newblock Springer series in statistics New York, NY, USA:.

\bibitem[George E. P.~Box, 1973]{BoxTiao1973}
George E. P.~Box, G. C.~T. (1973).
\newblock {\em Bayesian Inference in Statistical Analysis (Wiley Classics
  Library)}.
\newblock Wiley-Interscience.

\bibitem[Hassibi and Stork, 1993]{hassibi1993second}
Hassibi, B. and Stork, D.~G. (1993).
\newblock Second order derivatives for network pruning: Optimal brain surgeon.
\newblock In {\em Advances in neural information processing systems}, pages
  164--171.

\bibitem[Hertz et~al., 2013]{hertz2013principles}
Hertz, J., Roudi, Y., Tyrcha, J., Quiroga, R.~Q., and Panzeri, S. (2013).
\newblock Principles of neural coding.

\bibitem[Ibrahim and Laud, 1991]{ibrahim1991bayesian}
Ibrahim, J.~G. and Laud, P.~W. (1991).
\newblock On bayesian analysis of generalized linear models using jeffreys's
  prior.
\newblock {\em Journal of the American Statistical Association},
  86(416):981--986.

\bibitem[Jeffreys, 1946]{Jeffreys46}
Jeffreys, H. (1946).
\newblock An invariant form for the prior probability in estimation problems.
\newblock {\em Proceedings of the Royal Society of London A: Mathematical,
  Physical and Engineering Sciences}, 186(1007):453--461.

\bibitem[Kass and Wasserman, 1996]{kass1996selection}
Kass, R.~E. and Wasserman, L. (1996).
\newblock The selection of prior distributions by formal rules.
\newblock {\em Journal of the American Statistical Association},
  91(435):1343--1370.

\bibitem[Koh et~al., 2007]{Kim07}
Koh, K., Kim, S.-J., and Boyd, S. (2007).
\newblock {An Interior-Point Method for Large-Scale l1-Regularized Logistic
  Regression}.
\newblock {\em Journal of Machine Learning Research}, (8):1519--1555.

\bibitem[LaMont and Wiggins, 2017]{lamont2017correspondence}
LaMont, C.~H. and Wiggins, P.~A. (2017).
\newblock A correspondence between thermodynamics and inference.
\newblock {\em arXiv preprint arXiv:1706.01428}.

\bibitem[LeCun et~al., 1990]{lecun1990optimal}
LeCun, Y., Denker, J.~S., and Solla, S.~A. (1990).
\newblock Optimal brain damage.
\newblock In {\em Advances in neural information processing systems}, pages
  598--605.

\bibitem[Lichtman, 2016]{Lichtman2016}
Lichtman, A.~J. (2016).
\newblock The keys to the white house: The current forecast for 2016.
\newblock {\em Social Education}, 80(1):26--30.

\bibitem[Lichtman and Keilis-Borok, 1981]{Lichtman81}
Lichtman, A.~J. and Keilis-Borok, V.~I. (1981).
\newblock Pattern recognition applied to presidential elections in the united
  states, 1860-1980: Role of integral social, economic, and political traits.
\newblock {\em Proceedings of the National Academy of Sciences},
  78(11):7230--7234.

\bibitem[Marre et~al., 2009]{marre2009prediction}
Marre, O., El~Boustani, S., Fr{\'e}gnac, Y., and Destexhe, A. (2009).
\newblock Prediction of spatiotemporal patterns of neural activity from
  pairwise correlations.
\newblock {\em Physical review letters}, 102(13):138101.

\bibitem[Mattingly et~al., 2018]{mattingly2018maximizing}
Mattingly, H.~H., Transtrum, M.~K., Abbott, M.~C., and Machta, B.~B. (2018).
\newblock Maximizing the information learned from finite data selects a simple
  model.
\newblock {\em Proceedings of the National Academy of Sciences},
  115(8):1760--1765.

\bibitem[Myung et~al., 2000]{Myung00}
Myung, I.~J., Balasubramanian, V., and Pitt, M.~A. (2000).
\newblock Counting probability distributions: Differential geometry and model
  selection.
\newblock {\em Proceedings of the National Academy of Sciences},
  97(21):11170--11175.

\bibitem[Nelder and Wedderburn, 1972]{Nelder1972}
Nelder, J.~A. and Wedderburn, R. W.~M. (1972).
\newblock Generalized linear models.
\newblock {\em Journal of the Royal Statistical Society. Series A (General)},
  135(3):370--384.

\bibitem[Ravikumar et~al., 2010]{ravikuumar10}
Ravikumar, P., Wainwright, M.~J., and Lafferty, J.~D. (2010).
\newblock High-dimensional ising model selection using l1-regularized logistic
  regression.
\newblock {\em Ann. Statist.}, 38(3):1287--1319.

\bibitem[Rissanen, 1987]{Rissanen1987}
Rissanen, J. (1987).
\newblock Stochastic complexity.
\newblock {\em Journal of the Royal Statistical Society. Series B
  (Methodological)}, 49(3):223--239.

\bibitem[Rissanen, 1996]{Rissanen1996}
Rissanen, J.~J. (1996).
\newblock Fisher information and stochastic complexity.
\newblock {\em IEEE Transactions on Information Theory}, 42(1):40--47.

\bibitem[Roudi et~al., 2015]{roudi2015multi}
Roudi, Y., Dunn, B., and Hertz, J. (2015).
\newblock Multi-neuronal activity and functional connectivity in cell
  assemblies.
\newblock {\em Current opinion in neurobiology}, 32:38--44.

\bibitem[Roudi and Hertz, 2011]{Yasser11}
Roudi, Y. and Hertz, J. (2011).
\newblock Mean field theory for nonequilibrium network reconstruction.
\newblock {\em Phys. Rev. Lett.}, 106:048702.

\bibitem[Saeys et~al., 2007]{saeys2007review}
Saeys, Y., Inza, I., and Larra{\~n}aga, P. (2007).
\newblock A review of feature selection techniques in bioinformatics.
\newblock {\em bioinformatics}, 23(19):2507--2517.

\bibitem[Schneidman et~al., 2006]{Schneidman06}
Schneidman, E., Berry, M.~J., Segev, R., and Bialek, W. (2006).
\newblock Weak pairwise correlations imply strongly correlated network states
  in a neural population.
\newblock {\em Nature}, 440(7087):1007--1012.

\bibitem[Schwarz, 1978]{BIC}
Schwarz, G. (1978).
\newblock Estimating the dimension of a model.
\newblock {\em Ann. Statist.}, 6(2):461--464.

\bibitem[Tibshirani, 1996]{Lasso}
Tibshirani, R. (1996).
\newblock Regression shrinkage and selection via the lasso.
\newblock {\em Journal of the Royal Statistical Society (Series B)},
  58:267--288.

\bibitem[Truett et~al., 1967]{truett1967multivariate}
Truett, J., Cornfield, J., and Kannel, W. (1967).
\newblock A multivariate analysis of the risk of coronary heart disease in
  framingham.
\newblock {\em Journal of Clinical Epidemiology}, 20(7):511--524.

\end{thebibliography}

\newpage

\appendix

\section{Upper bound for models including the bias term}\label{sec:field}
Here we prove that due to the parity of the hyperbolic cosine, the complexity of a logistic regression model with $n$ binary inputs, where $n-1$ of them are uniformly distributed, is equal to the complexity of the same model when all $n$ inputs are uniformly distributed, which achieves the upper bound, as we found in section \ref{sec:mathematics}. 

To prove the statement, let's consider the Fisher Information elements for a model with $n$ inputs connected through a $n$ dimensional vector of parameters $\bm\theta$
\begin{equation}
F_{i,j}(\bm{\theta}) = \sum_{\bm{x}} \nu(\bm{x}) \cosh^{-2}(\bm{\theta}\cdot\bm{x})x_i x_j.
\end{equation}
Let's perform the sum over an arbitrary variable $x_k$,
\begin{equation}\label{eqn:rewritingFisher1}
\begin{array}{lll}
F_{i,j}(\bm{\theta}) &= &\displaystyle\sum_{\bm{x}_{/k}} \nu^{+}(\bm{x}_{/k}) \cosh^{-2}(\theta_k + \bm{\theta}_{/k}\cdot\bm{x}_{/k})x_i x_j \, + \\
&+ &\displaystyle\sum_{\bm{x}_{/k}} \nu^{-}(\bm{x}_{/k}) \cosh^{-2}(\theta_k - \bm{\theta}_{/k}\cdot\bm{x}_{/k})x_i x_j,
\end{array}
\end{equation}
where $k \neq i,j$, $\bm{x}_{/k}$ and $\bm{\theta}_{/k}$ are the $n-1$ dimensional vectors obtained by removing the $k$-th index and $\nu^{+}(\bm{x}_{/k}) = \nu(x_k = 1;\bm{x}_{/k})$ and $\nu^{-}(\bm{x}_{/k}) = \nu(x_k = -1;\bm{x}_{/k})$. By changing $\bm{x}_{/k}$ with $-\bm{x}_{/k}$ in the second summation we obtain
\begin{equation}\label{eqn:rewritingFisher2}
\begin{array}{lll}
F_{i,j}(\bm{\theta}) &= &\displaystyle\sum_{\bm{x}_{/k}} \nu^{+}(\bm{x}_{/k}) \cosh^{-2}(\theta_k + \bm{\theta}_{/k}\cdot\bm{x}_{/k})x_i x_j \, + \\
&+ &\displaystyle\sum_{-\bm{x}_{/k}} \nu^{-}(-\bm{x}_{/k}) \cosh^{-2}(\theta_k + \bm{\theta}_{/k}\cdot\bm{x}_{/k})x_i x_j,
\end{array}
\end{equation}
Notice that $\sum_{\bm{x}_{/k}}$ and $\sum_{-\bm{x}_{/k}}$ is a sum over the same elements but in a reversed order. It follows that we can write it as
\begin{equation}\label{eqn:rewritingFisher3}
F_{i,j}(\bm{\theta}) = \sum_{\bm{x}_{/k}} \left(\nu^{+}(\bm{x}_{/k})+\nu^{-}(-\bm{x}_{/k})\right) \cosh^{-2}(\theta_k + \bm{\theta}_{/k}\cdot\bm{x}_{/k})x_i x_j.
\end{equation}
We have considered here $k\neq i,j$, but the latter equation is true for any elements of the Fisher Information matrix. In fact, it is not difficult to see that equation \ref{eqn:rewritingFisher3} holds in the case when $k=i=j$. For the cases $k=i\neq j$ and $k=j\neq i$, there will be a minus sign in front of the second summation in equation \ref{eqn:rewritingFisher1} which will become again a plus sign in equation \ref{eqn:rewritingFisher2} after transforming $\bm{x}_{/k}$ in $-\bm{x}_{/k}$. Now, if the distribution of the inputs satisfy the relation
\begin{equation}\label{eqn:relation}
\nu(x_k = 1;\bm{x}_{/k})+\nu(x_k = -1;-\bm{x}_{/k}) = \nu(\bm{x}_{/k})
\end{equation}
then the elements of the Fisher Information matrix can be written as if the k-th input was absent and the corresponding weight was the bias term acting on the output, namely
\begin{equation}\label{eqn:final}
F_{i,j}(\bm{\theta}) = \sum_{\bm{x}_{/k}} \nu(\bm{x}_{/k}) \cosh^{-2}(\theta_k + \bm{\theta}_{/k}\cdot\bm{x}_{/k})x_i x_j.
\end{equation}
In particular, equation \ref{eqn:relation} is true if the distribution factorize over the spin variable $k$, i.e. $\nu(\bm{x}) = \nu(x_k)\nu(\bm{x}_{/k})$ and if all possible patterns and anti-patterns of the remaining $n-1$ spin are observed with the same frequency, i.e. $\nu(\bm{x}_{/k}) = \nu(-\bm{x}_{/k})$. One instance in which this is true is when $n-1$ out of the $n$ inputs are uniformly distributed, i.e. $\nu(\bm{x}_{/k}) = 2^{-n+1}$, regardless of $\nu(x_k)$. It follows that, the upper bound for the complexity is reached already in the case in which the input distribution is flat over the configuration space of only $n-1$ inputs, as it has been observed in the simulations and discussed in the paper in section \ref{sec:mathematics}.

As a consequence, since the bias term can be thought as a weight parameter connecting a constant input (always equal to $1$ or $-1$), the complexity of a logistic model with $n-1$ inputs (weights) and the bias term reaches the same upper bound as the model with $n$ inputs (weights) and no bias. The upper bound is attained when the $n-1$ inputs are uniformly distributed.

\section{Proof of equation \ref{eqn:regularized}}\label{sec:proof}
We consider logistic models with parameters defined on a finite support $|\theta_k| \leq \Theta$, $\forall k$ with $\Theta$ being any finite positive number, in the case of uniformly distributed inputs, $\nu(\bm{x}) = 2^{-n}$, $\forall \bm{x}$. Recall that the Fisher Information matrix elements are given by $F_{i,j}(\bm{\theta}) = \sum_{\bm{x}} \nu(\bm{x}) \cosh^{-2}(\bm{\theta}\cdot\bm{x})x_i x_j$.

We examine first the off-diagonal terms. We carry out the summations over the variables $i$ and $j$ and define the variables $\hat{\bm\theta}$ and $\bm{\hat{x}}$ by removing the indices $i$ and $j$ from the variables $\bm\theta$ and $\bm{x}$ respectively,
\begin{equation}
\begin{array}{lll}
F_{i,j} &=& 2^{-n} \sum_{\hat{\bm{x}}} \left[\cosh^{-2}(\theta_i + \theta_j+\hat{\bm\theta}\cdot\bm{\hat{x}}) - \cosh^{-2}(\theta_i - \theta_j+ \hat{\bm\theta}\cdot\bm{\hat{x}}) + \right. \\
&-& \left.\cosh^{-2}(-\theta_i + \theta_j+\hat{\bm\theta}\cdot\bm{\hat{x}}) + \cosh^{-2}(-\theta_i - \theta_j+\hat{\bm\theta}\cdot\bm{\hat{x}})\right], \\
&=& 2^{-n} \sum_{\hat{\bm{x}}} \left[\cosh^{-2}(\theta_i + \theta_j+\hat{\bm\theta}\cdot\bm{\hat{x}}) - \cosh^{-2}(\theta_i - \theta_j+ \hat{\bm\theta}\cdot\bm{\hat{x}}) + \right. \\
&-& \left.\cosh^{-2}(\theta_i - \theta_j - \hat{\bm\theta}\cdot\bm{\hat{x}}) + \cosh^{-2}(\theta_i + \theta_j-\hat{\bm\theta}\cdot\bm{\hat{x}})\right]\\
&=& 2^{-n+1} \sum_{\hat{\bm{x}}} \left[\cosh^{-2}(\theta_i + \theta_j+\hat{\bm\theta}\cdot\bm{\hat{x}}) - \cosh^{-2}(\theta_i - \theta_j+ \hat{\bm\theta}\cdot\bm{\hat{x}})\right],
\end{array}
\end{equation}
where we have first exploited the parity of the hyperbolic cosine and then rearranged the terms in the summations. We think of the average over observations as the average over a random process with the same limiting distribution, i.e. $ p(\bm{x}) = \nu(\bm{x})$. This will allow us to exploit central limit theorem and perform calculations. It follows that, since the distribution over $\bm{x}$ is uniform, the quantity $z \equiv \hat{\bm\theta}\cdot\bm{\hat{x}} = \sum_{k\neq i,j} \theta_k x_k$ is the sum of $n-2$ independent random variables, $z_k = \theta_k x_k$, distributed as $p(z_k = \pm \theta_k) = 1/2$. Therefore each random variable $z_k$ has zero mean and variance $\theta_k^2$. Asymptotically, i.e. as $n$ goes to infinity, $z$ becomes a Gaussian random variable with zero mean and variance $r^2 = \sum_{k\neq i,j} \theta_k^2$. This is ensured by Lyapunov's condition which is always satisfied when the parametric family is bounded \citep[see for instance][]{Billingsley95}. Thus, in the last equation, we substitute the expectation over the configuration states $\hat{\bm{x}}$ with an expectation over $z$, 
\begin{equation}\label{eqn:asymptotic_offdiagonal}
\begin{array}{lll}
F_{i,j}(\bm{\theta}) &=& \frac{1}{2}\int_{-\infty}^{\infty} dz\, p(z) \left[\cosh^{-2}(\theta_i + \theta_j + z) - \cosh^{-2}(\theta_i - \theta_j + z)\right] = \\
&=& \frac{1}{2}\int_{-\infty}^{\infty} dz\, p(z-\theta_i) \left[\cosh^{-2}(z + \theta_j) - \cosh^{-2}(z- \theta_j)\right]   
\end{array}
\end{equation}
where $p(z) = e^{-z^2/2 r^2}/\sqrt{2\pi r^2}$ is the gaussian density and where we used the change of variable $z \rightarrow z+\theta_i$ for passing from the first to the second line. 
The integrand in the last equation is the difference between two functions centered at $z=\pm\theta_j$ and weighted according to a gaussian density centered at $z=\theta_i$. 
However the standard deviation of the Gaussian density grows with $n$ and eventually for $n \gg 4\Theta^2$, the two contributions will be weighted equally cancelling each other. More specifically, for $r \gg 2\Theta$, namely when $n \gg 4\Theta^2$, the density can be approximated as $p(z) \sim 1/\sqrt{2\pi r^2} + O(1/r^3)$. The term in $O(1/r)$ has a zero contribution to the integral so the first non zero contribution comes from the $O(1/r^3)$ term. Thus off-diagonal elements go to zero faster than $O(n^{-1/2})$ which ensures that these terms will not bring a relevant contribution to the determinant as $n$ goes to infinity.

Following the same argument, it is easy to obtain a simple expression also for the diagonal terms $F_{i,i}(\bm{\theta}) = \sum_{\mu} \nu(\bm{x}^\mu) \cosh^{-2}(\bm{\theta}\cdot\bm{x}^\mu)$, $\forall i$. The latter will become $F_{i,i}(\bm{\theta}) = \int_{-\infty}^{\infty} dz\, p(z) \cosh^{-2}(z)$ where $p(z) = e^{-z^2/2 r^2}/\sqrt{2\pi r^2}$ and $r^2 = \sum_k \theta_k^2$. The elements on the diagonal are all equal and their value depends only on $r$, namely $F_{i,i}(\bm{\theta}) = \lambda(r)$ where
\begin{equation}
\lambda(r) = \int_{-\infty}^{\infty} dz\, \frac{e^{-z^2/2 r^2}}{\sqrt{2\pi r^2}} \cosh^{-2}(z).
\end{equation}
The square root of the determinant of the Fisher Information is simply given by $\lambda^{n/2}(r)$ and, given that $\lambda(r)$ depends only on the euclidean distance $r$ in the space of parameters, it's easier to perform the integral in spherical coordinates. Thus, $e^C = \int d\bm\theta \lambda^{n/2}(r(\bm\theta))$ becomes $e^C = \int d\Omega dr \,r^{n-1}\lambda^{n/2}(r)$
and by integrating over the solid angle $\Omega$, it becomes
\begin{equation}
e^C = \frac{2\pi^{n/2}}{\Gamma(n/2)}\int_{0}^{\sqrt{n}\Theta} dr\, r^{n-1} \lambda^{n/2}(r),
\end{equation}
where $\Gamma(x)$ is the Euler gamma function. In the large $r$ limit, since $\lambda(r) \sim \sqrt{2/\pi r^2} +O(1/r^3)$, the product $r^{n-1} \lambda(r)^{n/2} \sim ({2/\pi})^{n/4} r^{n/2-1}$. Employing the latter approximation, the above integral becomes $e^C \sim 2(2\pi\Theta^2 n)^{n/4}/\Gamma(n/2+1)$. Finally, using Stirling's approximation for the gamma function, $\Gamma(n/2+1) \sim \sqrt{\pi n} (n/2e)^{n/2}$, the complexity in the large $n$ regime can be approximated by equation \ref{eqn:regularized}
\begin{equation}\label{eqn:asymptotic_expression}
e^C \sim \frac{2}{\sqrt{\pi n}}\left(\frac{8\pi\Theta^2 e^2}{n}\right)^{n/4}.
\end{equation}

\end{document}